\documentclass[10pt,twocolumn,letterpaper]{article}

\usepackage{cvpr}      
\usepackage{indentfirst}
\usepackage[dvipsnames]{xcolor}

\definecolor{cvprblue}{rgb}{0.21,0.49,0.74}
\usepackage[pagebackref,breaklinks,colorlinks,citecolor=cvprblue]{hyperref}

\def\paperID{12683} 
\def\confName{CVPR}
\def\confYear{2024}

\title{Solving the long-tailed distribution problem by exploiting the synergies and balance of different techniques}

\author{Ziheng Wang, Toni Lassila, Sharib Ali\\
Faculty of Engineering and Physical Sciences, University of Leeds, Leeds, LS2 9JT, UK\\
{\tt\small Ziheng-Wang@outlook.com, T.Lassila@leeds.ac.uk, S.S.Ali@leeds.ac.uk}
}

\begin{document}
\maketitle
\begin{abstract}
In real-world data, long-tailed data distribution is common, making it challenging for models trained on empirical risk minimization to learn and classify tail classes effectively. While many studies have sought to improve long-tail recognition by altering the data distribution in the feature space and adjusting model decision boundaries, research on the synergy and corrective approach among various methods is limited. Our study delves into three long-tail recognition techniques: Supervised Contrastive Learning (SCL), Rare-Class Sample Generator (RSG), and Label-Distribution-Aware Margin Loss (LDAM). SCL enhances intra-class clusters based on feature similarity and promotes clear inter-class separability, but tends to favor dominant classes only. When RSG is integrated into the model, we observed that the intra-class features further cluster towards the class center, which demonstrates a synergistic effect together with SCL's principle of enhancing intra-class clustering. RSG generates new tail features and compensates for the tail feature space squeezed by SCL. Similarly, LDAM is known to introduce a larger margin specifically for tail classes; we demonstrate that LDAM further bolsters the model's performance on tail classes when combined with the more explicit decision boundaries achieved by SCL and RSG. Furthermore, SCL can compensate for the dominant class accuracy sacrificed by RSG and LDAM. Our research emphasizes the synergy and balance among the three techniques, with each amplifying the strengths of the others and mitigating their shortcomings. Our experiment on long-tailed distribution datasets, using an end-to-end architecture, yields competitive results by enhancing tail class accuracy without compromising dominant class performance, achieving a balanced improvement across all classes.
\end{abstract}

\section{Introduction}
\label{sec:intro}
\begin{figure*}
  \centering
  
  \begin{subfigure}{0.3\linewidth}
    \includegraphics[width=\linewidth]{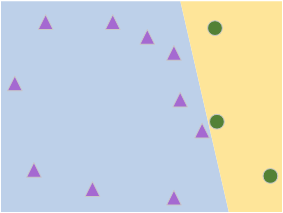}
    \caption{}
    \label{fig:CE_feature_space}
  \end{subfigure}
  \hfill
  \begin{subfigure}{0.3\linewidth}
    \includegraphics[width=\linewidth]{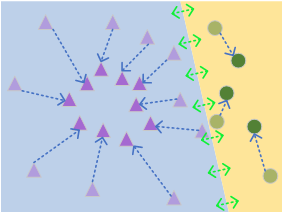}
    \caption{}
    \label{fig:SCL_process}
  \end{subfigure}
  \hfill
  \begin{subfigure}{0.3\linewidth}
    \includegraphics[width=\linewidth]{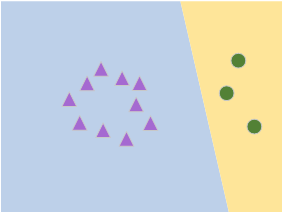}
    \caption{}
    \label{fig:SCL_results}
  \end{subfigure}
  
  \begin{subfigure}{0.3\linewidth}
    \includegraphics[width=\linewidth]{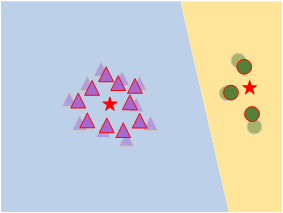}
    \caption{}
    \label{fig:CESC}
  \end{subfigure}
  \hfill
  \begin{subfigure}{0.3\linewidth}
    \includegraphics[width=\linewidth]{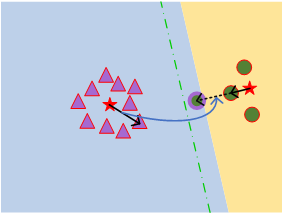}
    \caption{}
    \label{fig:MV}
  \end{subfigure}
  \hfill
  \begin{subfigure}{0.3\linewidth}
    \includegraphics[width=\linewidth]{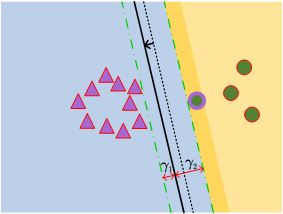}
    \caption{}
    \label{fig:LDAM}
  \end{subfigure}
  \caption{(a) shows the feature distribution for long-tail recognition based on Cross-Entropy (CE), where purple triangles represent head classes and green circles represent tail classes~\cite{Hybrid_contrastive}. (b) Supervised Contrastive Learning (SCL) in binary classification promotes intra-class clustering and inter-class separation, which can be regarded as a mutual repulsion (represented by green arrows) between the clusters of the two distinct classes. The shadow represents the position of the feature under CE classification, and the blue arrows indicate the trajectory of feature movement. (c) Feature distribution under the influence of SCL. Ideally, where both classes are tightly clustered, a void feature region exists between them, devoid of any feature distribution, signifying a clear separation of classes. (d) Rare-Class Sample Generator (RSG) calculates a set of class centers, and for clarity, only one class center (red star) is drawn for each class. RSG drives the features of a class towards its class center. Features with a red outline further cluster towards the class center from their position under SCL (shadowed area). (e) RSG transfers the feature displacement (black arrow) between the head class samples and the class center to tail samples, generating a new tail sample (green circle with a purple outline). RSG expands the feature space of tail samples, thereby influencing the potential decision boundary. (f) Label-Distribution-Aware Margin Loss (LDAM) calculates the margin based on the number of samples of each class, providing a larger margin, $\gamma_2$, for tail samples. The solid black line represents the adjusted decision boundary by LDAM.}
  \label{fig:feature_space}
\end{figure*}

\noindent{Computer vision} has achieved significant successes and has been widely used in real-world applications due to advancements in deep convolutional neural networks (CNNs) and the availability of large and high-quality datasets. However, real-world data, in general, follows a long-tailed distribution. As a result, CNN models trained on balanced datasets struggle to perform well on skewed datasets, especially for classes with few samples (tail classes). These tail classes are of imminent importance as they usually represent critical problems that could widely benefit society, such as severe diseases, endangered species, fraud activities, etc. Therefore, training unbiased models for long-tail recognition (LTR) is of great practical significance.

Traditional re-sampling and re-weighting methods fail to address LTR well \cite{Hybrid_contrastive,Parametric_contrastive,BBN,RIDE,Self-Heterogeneous_Integration_with_Knowledge_Excavation,C2am,Long-tailed_recognition_via_weight_balancing}. Thus, some logit adjustment approaches \cite{logit_adjustment,LDAM,Good_and_Bad} that directly adjust decision boundaries for a balanced feature distribution have been proposed and widely researched. Recent studies have improved classification accuracy using ensemble models~\cite{BBN,OLTR,TADE}, but several studies have often introduced greater complexity. For example, decoupled learning~\cite{LDAM,Supervised,Targeted_contrastive,BBN} adopts a two-stage approach, focusing on representation learning and classifier training separately to optimize each without conflicts.

Supervised Contrastive Learning (SCL), as detailed in \cite{Supervised}, offers robust capabilities for representation learning on balanced datasets. Originally designed for two-stage training, SCL's potential in LTR is further unlocked in an end-to-end hybrid network, Hybrid-SC in \cite{Hybrid_contrastive}. The approach combines SCL for feature learning with Cross-Entropy (CE) for classification, outperforming the conventional two-stage SCL method only. The Hybrid-SC highlights the effectiveness of SCL in long-tail recognition, and the same authors also designed a prototypical supervised contrastive learning strategy (Hybrid-PSC) to save memory space. The Balanced Contrastive Learning (BCL)~\cite{Balanced_contrastive} extends SCL by integrating Class-averaging and Class-complement strategies to ensure equitable learning across both head and tail classes. The BCL leverages data augmentation to enhance representation learning. Additionally, it employs a logit adjustment classifier~\cite{logit_adjustment} to bias the model towards tail classes further. The state-of-the-art (SOTA) results demonstrate the efficacy and synergy of logit adjustment classification with SCL-based representation learning in LTR. 


Our method is based on principles from multi-task learning~\cite{Multi-task}, an end-to-end model with distinct branches to optimize SCL, Rare-Class Sample Generator (RSG) \cite{RSG}, and Label-Distribution-Aware Margin Loss (LDAM) \cite{LDAM} individually. SCL's performance is inherently linked to class sample sizes, favoring head classes \cite{Parametric_contrastive, Balanced_contrastive} and thus limiting the representation of tail classes. In contrast, LDAM has been shown to significantly enhance the accuracy of tail classes, more so than other loss functions, at the cost of head class accuracy \cite{survey}. The integration of RSG into the model exacerbates this effect, further shifting the model attention towards tail classes\cite{RSG}. Our goal is to achieve a synergy where each component capitalizes on its strengths to compensate for the limitations of the others. While SCL tends to compress tail classes into tighter feature spaces, RSG works to expand these spaces, as shown in Figure \ref{fig:MV}. SCL and RSG promote more distinct class clustering with enhanced inter-class separation, as illustrated in Figures \ref{fig:SCL_process} to \ref{fig:CESC}, mitigating the negative impact on head classes when LDAM is applied to introduce larger margins for tail classes, as demonstrated in Figure \ref{fig:LDAM}.

Our main contributions are as follows:
\begin{itemize}
    \item In our study, by simply minimizing a weighted linear combination of losses, we identified a synergistic and compensatory relationship between SCL, RSG, and LDAM for long-tail recognition, which yielded competitive results.
    \item Ablation experiments revealed that balancing the strengths and weaknesses of SCL, RSG, and LDAM offers an effective solution for long-tail recognition challenges.
    \item We emphasize that when classes are distinctly separable, expanding the feature space for tail classes to boost their accuracy does not compromise the performance of other classes.
\end{itemize}


\section{Related work}
\label{sec:related_work}

\begin{figure*}[t]
  \centering
   \includegraphics[width=0.9\linewidth]{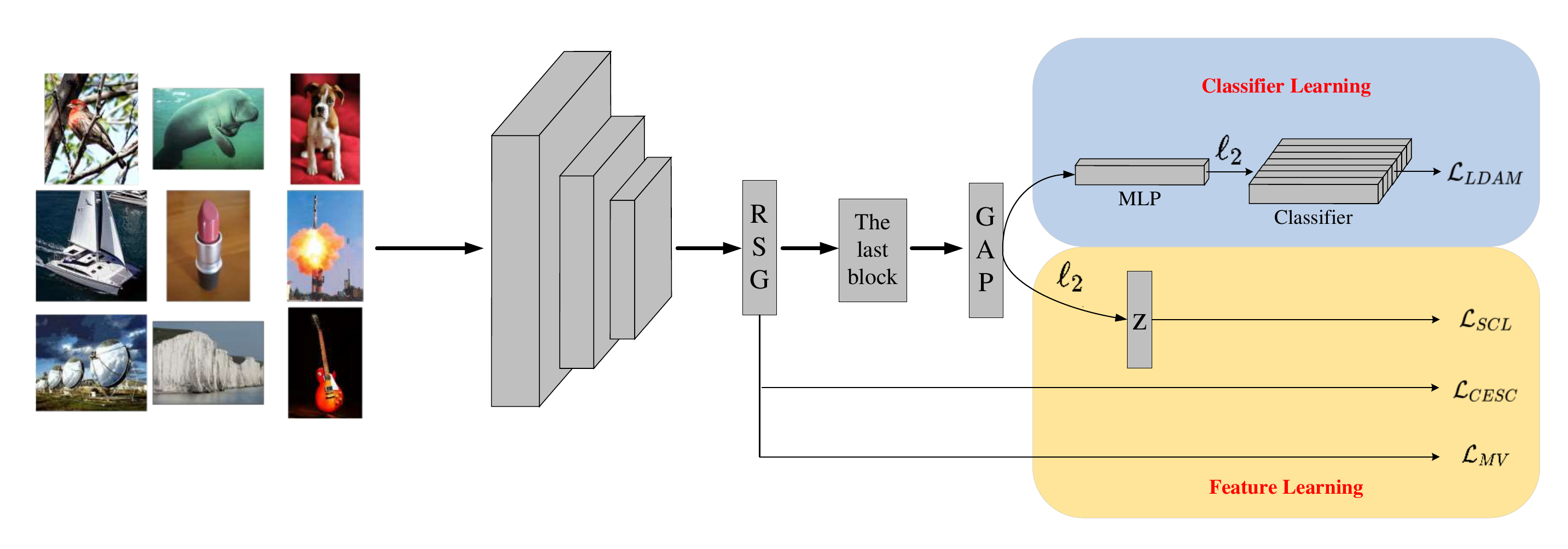}
   \caption{Our model overview diagram. RSG is the abbreviation of the Rare-Class Sample Generator. Based on the experimental results from \cite{RSG}, it is typically integrated before the last block of the ResNet \cite{resnet} model. GAP refers to Global Average Pooling.}
   \label{fig:model_structure}
\end{figure*}



\noindent\textbf{Long-tailed Recognition.} Re-balancing encompasses two main techniques: data re-sampling and loss re-weighting. When under-sampling head classes, valuable information can be lost, leading to a reduction in the model's generalization capability and an obstruction in feature learning 
\cite{Hybrid_contrastive,Reslt,BBN,Balanced_Product_of_Calibrated_Experts}. Over-sampling tail classes can result in model over-fitting \cite{Hybrid_contrastive,Reslt,BBN,Balanced_Product_of_Calibrated_Experts} because the features of the tail class are not truly enriched. Loss re-weighting seeks to adjust the loss values of different classes during training via the inverse frequency of class samples \cite{Focal}, the number of effective samples \cite{Class_Balance_Loss}, etc. However, the re-weighting method has been proved to complicate optimization \cite{hard_to_optimize_1,hard_to_optimize_2}, especially in large-scale datasets. Research \cite{BBN,Decoupling_representation_and_classifier} has revealed that re-balancing strategies are beneficial for classification learning but counterproductive for representation learning. Therefore, in \cite{LDAM} the first stage is dedicated to representation learning, and the second stage employs re-weighting strategies to fine-tune the classifier. The two-stage methodology has been successful in achieving superior accuracy \cite{No_One_Left_Behind,Long-tailed_recognition_via_weight_balancing} but its inference speed is lower than the end-to-end architecture \cite{adaptive}. Gradients of decoupling network cannot backpropagate from the top layers to the bottom layers of CNNs \cite{RSG}, and it are sensitive to variations in hyperparameters and optimizers \cite{Decoupled_contrastive_learning}.  Ensemble learning \cite{BBN,OLTR,TADE}, which leverages the expertise of multiple models to learn varied knowledge, enhances accuracy. However, it's necessary to trade off the increased complexity and computational resources \cite{Trustworthy_long-tailed_classification,Nested}.

\noindent\textbf{Adjusting Feature Distribution and Decision Boundaries} Due to limited sample sizes of tail class, its feature space is constricted to a narrow region \cite{Balancing_Logit_Variation}. Several studies \cite{gaussian_clouded_logit_adjustment,RSG} have focused on expanding the feature space of these tail classes. By generating new tail samples, \cite{RSG} not only expands tail class feature space but also ensures that class features gravitate closer to the class center. Supervised Contrastive Learning \cite{Supervised,Hybrid_contrastive,Balanced_contrastive,Targeted_contrastive,Parametric_contrastive} drives that samples of the same class are closely clustered while those from different classes are spread out, thereby achieving clear boundaries between classes in the feature space. Besides, logits adjustment \cite{logit_adjustment,LDAM} introduces the margin to adjust the decision boundary of the model. It provides the tail class with a larger margin to offset the bias caused by imbalanced datasets. 



\section{Adjusting the Feature Distribution and Decision Boundary for Long-tailed Learning}
\subsection{Preliminaries}

\noindent The essence of deep learning is to find the optimal mapping function $\varphi$  from input $\mathcal{X}$ to output $\mathcal{Y}$ through a fitting process. The function $\varphi$ typically consists of two parts: a nonlinear encoder $f$: $\mathcal{X}\to  \mathcal{Z}\in \mathbb{R}^h$ and a linear classifier $W$: $\mathcal{Z} \to \mathcal{Y}$, resulting in $ \varphi(x) = W \cdot f(x) + b$. High-quality features $\mathcal{Z}$ provide the classifier with valuable information for achieving high accuracy. The performance of the classifier also directly impacts the model's overall effectiveness. Thus, for long-tail recognition tasks, our goal is to balance the strengths and weaknesses of a high-quality encoder and an effective classifier to achieve superior results. Next, we recall the key concepts utilized in our research to facilitate subsequent understanding and application.

\textbf{Supervised Contrastive Loss} For a sample $x_i$ of class $y$ in a batch $B$, its feature representation is denoted as $z_i$. Following the approach in \cite{Balanced_contrastive}, the formulation of Supervised Contrastive Loss (SCL) for class $y$ within the batch is as follows.

\begin{equation}
    \mathcal{L}_{SCL}^y= \sum_{i\in B_y}^{}-\frac{1}{|B_{y}|-1}\sum_{p\in B_{y}\setminus\{i\}}\log\frac{\exp(z_{i}\cdot z_{p}/\tau)}{\displaystyle\sum_{k\in B\setminus\{i\}}\exp(z_{i}\cdot z_{k}/\tau)}
    \label{eq:SCL}
\end{equation}

where $B_y$ represents all samples of class $y$ in batch $B$, and $\left | \cdot \right | $ denotes the number of samples in the set. $\tau$ is the scaling temperature hyperparameter.

\textbf{Rare-Class Sample Generator} The RSG \cite{RSG} comprises three modules: the Center Estimation Module, the Contrastive Module, and the Vector Transformation Module, as well as two loss functions: the Center Estimation with Sample Contrastive Loss (CESC) and the Maximized Vector Loss (MV). 

The Center Estimation Module estimates a set of class centers for each class to adapt to the multimodal distribution of the data. For inputs belonging to class $y$, the module outputs a probability distribution $\gamma^y$, which indicates the likelihood of the input being associated with each of the centers within class $y$. The Contrastive Module randomly selects two samples from a mini-batch and calculates the probability $\gamma^{\ast }$, determining whether the two samples belong to the same class. The Vector Transformation Module aims to apply the feature displacement(fd) of frequent class features, $x_{\text{fd-freq}}$, to tail class samples. The feature displacement is defined as $x_{\text{fd-freq}}^y = x_{\text{freq}}^y - \text{up}(C_{\mathcal{K}}^y)$, where $x_{\text{freq}}^y$ is a sample from the frequent class, and $C_{\mathcal{K}}^y$ is the nearest center indexed by $\mathcal{K}$ to the sample $x_{\text{freq}}^y$ within class $y$. The function $\text{up}(\cdot)$ denotes upsampling of $C$ to align with the feature map of the sample. A linear transformation $\mathcal{T}$, given by $\mathcal{T}(z) = \text{conv}(z)$, generates a new tail sample $x_{\text{new}}^{y'} = \mathcal{T}(x_{\text{fd-freq}}^y) + x_{\text{rare}}^{y'}$.

The first term of the Center Estimation with Sample Contrastive (CESC) Loss aims to update class centers to cluster samples closer to their nearest class center. The second term is the Contrastive Module, which determines whether each pair of input samples comes from the same class. The formula for the CESC Loss, as referred from \cite{RSG}, is as follows:
\begin{align}
\mathcal{L}_{CECS} = & \left \langle \sum_{i=0}^{K-1} \gamma_i^y \sum_{c,w,h} \left \|  x_{(c,w,h)}^y - up(C_i^y)_{(c,w,h)}\right \|^2 \right \rangle_B \nonumber \\
& - \left \langle (t \log \gamma ^\ast + (1-t) \log(1-\gamma ^\ast )) \right \rangle_{\frac{B}{2} }.
\label{eq:CECS}
\end{align}

Here $c$, $w$, $h$ respectively denote the indices of the feature maps along the channel, width, and height dimensions. $\gamma_i^y$ represents the probability of the sample $x_{(c,w,h)}^y$ belonging to the $i$-th center. $K$ is the number of centers for class $y$, and $C$ denotes the class centers. $t \in \{0,1\}$ is the ground-truth label used to determine if samples belong to the same class. $\langle \cdot \rangle$ denotes the mean operation. The size of the mini-batch is consistent with the previous text as $B$. Since samples appear in pairs in the Contrastive Module, the second term is averaged via $B/2$.

The first term of the Maximum Vector (MV) Loss aims to align the transformed feature displacement of the frequent class, $\mathcal{T} (x_{fd-freq})$, with the feature displacement of the original rare class sample in the same direction. The second term ensures that the transformation $\mathcal{T}$ does not alter the original length of the feature displacement. Both the first and second terms impact cooperatively on the decision boundary by enlarging feature space of the rare class. The third term serves to eliminate class-related information from the frequent class. As referenced in \cite{RSG}, the MV Loss formula is as follows:

\begin{align}
& \mathcal{L}_{MV} =  \bigg\langle \sum_{w,h} \big( \big| \frac{\mathcal{T} (x_{fd-freq})^{(w,h)} \cdot x_{fd-rare}^{(w,h)}} {\big\| \mathcal{T} (x_{fd-freq})^{(w,h)} \big\|_2 \big\| x_{fd-rare}^{(w,h)} \big\|_2} -1 \big| \big) \bigg\rangle_{B_{n}} \nonumber\\
& + \bigg\langle \sum_{w,h} \big( \big| \big\| \mathcal{T} (x_{fd-freq})^{(w,h)} \big\|_2 - \big\| x_{fd-freq}^{(w,h)} \big\|_2 \big| \big) \bigg\rangle_{B_{n}} \nonumber\\
& - \big\langle \log \gamma^\ast \big\rangle_{B_{n}}
\end{align}

where \( w \) and \( h \) are the indices for width and height, respectively, \( |\cdot| \) denotes the absolute value, and \( \|\cdot\|_2 \) represents the L2 norm. \( x_{\text{fd-rare}} \) is the feature displacement obtained for a rare class sample from its nearest class center. \( B_n \) is the number of newly generated samples.

In RSG, an epoch threshold \( T_{\text{th}} \) is set. During the initial training phase (epoch \( \leq T_{\text{th}} \)), only \( \mathcal{L}_{\text{CESC}} \) is employed. As training progresses into the later phase (epoch \( > T_{\text{th}} \)), the second term (Contrastive Module) of \( \mathcal{L}_{\text{CESC}} \) stops updating, and model starts to generate new samples for tail classes, thereby activating \( \mathcal{L}_{\text{MV}} \).

\textbf{Label-Distribution-Aware Margin Loss}
According to the paper \cite{LDAM}, the form of the LDAM loss for an input sample $x_i$ with label $y_i$ and logit $z_i$ is as follows:

\begin{equation}
\mathcal{L}_{LDAM}(z_i,y_i)=-log\frac{e^{z_{i,y_i}-\Delta _{y_i}}}{e^{z_{i,y_i}-\Delta _{y_i}}+ {\textstyle \sum_{j\ne y_i}e^{z_{j,y} }} }
\label{eq:LDAM equation}
\end{equation}

After calculation and derivation of \cite{LDAM}, the margin $\Delta$ is concluded that for class $j$, $\Delta_j = \frac{C}{n_j^{1/4}}$. Here, $C$ is a hyperparameter that needs to be tuned, and $n_j$ is the number of samples of class $j$. The batch-wise LDAM loss can be formulated as follows:

\begin{equation}
    \mathcal{L}_{LDAM}=\frac1{|B|}\sum_{i\in B}\mathcal{L}_{LDAM}(z_i,y_i) 
\label{eq:B-LDAM equation}
\end{equation}

\subsection{Analysis and Methods}

\noindent \textbf{SCL and RSG} The \(\mathcal{L}_{\text{SCL}} \) encourages samples of the same class to cluster and keeps samples of different classes away from each other in the feature space, as depicted in Figure \ref{fig:SCL_process}. The second term of \(\mathcal{L}_{\text{CESC}} \) acts similarly with \(\mathcal{L}_{\text{SCL}} \) . The first term of \(\mathcal{L}_{\text{CESC}} \) promotes features clustering towards class centers, contributing to tighter class cohesion as shown in Figure \ref{fig:CESC}. The \(\mathcal{L}_{\text{MV}} \) of RSG generates tail samples to expand the feature space compressed by \(\mathcal{L}_{\text{SCL}} \), the process detailed in Figure \ref{fig:MV}, thus improving the model's ability to discern subtle differences in tail classes and enhancing correct classification probability. While \(\mathcal{L}_{\text{MV}} \) might increase intra-class variation for tail classes, new tail samples providing the additional positive pairs are beneficial for tail classes computing in \(\mathcal{L}_{\text{SCL}} \). The \(\mathcal{L}_{\text{SCL}} \) optimizes relative sample positions in the feature space to compensate for potential intra-class sample distribution diffusion caused by \(\mathcal{L}_{\text{MV}} \). Consequently, even as tail class samples occupy a larger area in the feature space, \(\mathcal{L}_{\text{SCL}} \) ensures intra-class clustering and inter-class separability. Hence, we think that a synergistic and compensatory relationship exists between SCL and RSG.

\noindent\textbf{SCL and LDAM} The \(\mathcal{L}_{\text{SCL}} \) and \(\mathcal{L}_{\text{LDAM}} \) have the same objective of enhancing inter-class separability. The \(\mathcal{L}_{\text{LDAM}} \) introduces larger margins for tail classes, which can reduce the accuracy of head classes due to misclassifying head samples near the decision boundary.  Ideally, when both head and tail classes are sufficiently clustered by \(\mathcal{L}_{\text{SCL}} \), a feature space void between classes emerges, as depicted in Figure \ref{fig:SCL_results}. In such scenarios, the classes are well-separated, which is conducive to decision boundary adjustment of \(\mathcal{L}_{\text{LDAM}} \), enhancing tail class accuracy while minimizing misclassification of head class samples. Thus, they can synergistically improve the model's classification performance across classes.

\noindent \textbf{RSG and LDAM} The relationship between \( \mathcal{L}_{\text{CESC}} \) of RSG and \(\mathcal{L}_{\text{LDAM}} \) is similiar with that between \(\mathcal{L}_{\text{SCL}} \) and \(\mathcal{L}_{\text{LDAM}} \). \(\mathcal{L}_{\text{MV}} \) of RSG, illustrated in Figure \ref{fig:MV}, potentially disperses the intra-class feature distribution to enlarge tail class feature space, aiding the model's ability to capture the diversity of tail classes. Concurrently, \(\mathcal{L}_{\text{LDAM}} \)  applies margins inversely proportional to class sample sizes, enhancing the model sensitivity to tail classes. The combination of \(\mathcal{L}_{\text{LDAM}} \) and \(\mathcal{L}_{\text{MV}} \)  is compensatory: \(\mathcal{L}_{\text{MV}} \)  enhances the representation of tail classes in feature space, while \(\mathcal{L}_{\text{LDAM}} \) ensures that, even within these expanded feature areas, the model maintains adequate margins for precise classification, as demonstrated in Figure \ref{fig:LDAM}. Following the aforementioned analysis, we posit that $\mathcal{L}_{SCL}$, $\mathcal{L}_{LDAM}$, $\mathcal{L}_{CESC}$, and $\mathcal{L}_{MV}$ exhibit both synergistic and compensatory relationships. Consequently, the loss function employed in our model training is as follows:
\begin{equation}
    \mathcal{L}_{total}=\alpha \mathcal{L}_{SCL}+\lambda \mathcal{L}_{LDAM}+\eta \mathcal{L}_{CESC}+\mu \mathcal{L}_{MV}
\label{eq:loss_total}
\end{equation}
\noindent where, $\alpha$, $\lambda$, $\eta$, and $\mu$ are hyperparameters to respectively adjust impact of four loss function.

\textbf{Framework} The overview of our model is depicted in Figure \ref{fig:model_structure}. Following the experiments in \cite{RSG}, we integrate RSG before the last block of ResNet \cite{resnet}. The $\mathcal{L}_{LDAM}$ is utilized for classifier learning, while $\mathcal{L}_{SCL}$, $\mathcal{L}_{CESC}$, and $\mathcal{L}_{MV}$ are employed for feature learning.
\label{sec:analysis_and_methods}
\section{Experiment}
\label{sec:experiment}

\begin{table*}[!h]
\centering
\begin{tabular}{l|ccc|ccc}

\hline
\bf{Method}                            & \multicolumn{3}{c|}{\bf{CIFAR-10-LT}}                                                           & \multicolumn{3}{c}{\bf{CIFAR-100-LT}}                                                         \\ \hline
Imbalance Factor, $\beta$                  & \multicolumn{1}{c|}{\bf 10}             & \multicolumn{1}{c|}{\bf 50}             & \bf 100            & \multicolumn{1}{c|}{\bf 10}            & \multicolumn{1}{c|}{\bf 50}             & \bf 100            \\ \hline \hline
CE$^\dagger$          & \multicolumn{1}{c|}{86.39}          & \multicolumn{1}{c|}{74.81}          & 70.36          & \multicolumn{1}{c|}{55.71}         & \multicolumn{1}{c|}{43.85}          & 38.32          \\ \hline
Focal loss$^\dagger$ \cite{Focal}                         & \multicolumn{1}{c|}{86.66}          & \multicolumn{1}{c|}{76.72}          & 70.38          & \multicolumn{1}{c|}{55.78}         & \multicolumn{1}{c|}{44.32}          & 38.41          \\ \hline
CB-Focal\cite{Class_Balance_Loss}                           & \multicolumn{1}{c|}{87.10}          & \multicolumn{1}{c|}{79.27}          & 74.57          & \multicolumn{1}{c|}{57.99}         & \multicolumn{1}{c|}{45.17}          & 39.60          \\ \hline
BBN\cite{BBN}                                & \multicolumn{1}{c|}{88.32}          & \multicolumn{1}{c|}{81.18}          & 79.82          & \multicolumn{1}{c|}{59.12}         & \multicolumn{1}{c|}{47.02}          & 42.56          \\ \hline
Causal model\cite{Good_and_Bad}                       & \multicolumn{1}{c|}{88.50}          & \multicolumn{1}{c|}{83.60}          & 80.60          & \multicolumn{1}{c|}{59.60}         & \multicolumn{1}{c|}{50.30}          & 44.10          \\ \hline
LDAM-DRW\cite{LDAM}                           & \multicolumn{1}{c|}{88.16}          & \multicolumn{1}{c|}{81.03}          & 77.03          & \multicolumn{1}{c|}{58.71}         & \multicolumn{1}{c|}{46.62}          & 42.04          \\ \hline
Hybrid-SC\cite{Hybrid_contrastive}                          & \multicolumn{1}{c|}{91.12}          & \multicolumn{1}{c|}{\underline{85.36}}          & \underline{81.40}          & \multicolumn{1}{c|}{63.05}         & \multicolumn{1}{c|}{51.87}          & 46.72          \\ \hline
MetaSAug-LDAM\cite{Metasaug}                      & \multicolumn{1}{c|}{89.68}          & \multicolumn{1}{c|}{84.34}          & 80.66          & \multicolumn{1}{c|}{61.28}         & \multicolumn{1}{c|}{52.27}          & 48.01          \\ \hline
BCL\cite{Balanced_contrastive}                                & \multicolumn{1}{c|}{91.12}          & \multicolumn{1}{c|}{\textbf{87.24}} & \textbf{84.32} & \multicolumn{1}{c|}{64.87}         & \multicolumn{1}{c|}{\textbf{56.59}} & \textbf{51.93} \\ \hline \hline

SCL-LDAM (ours)                      & \multicolumn{1}{c|}{\underline{91.18}} & \multicolumn{1}{c|}{83.40}          & 80.48          & \multicolumn{1}{c|}{\underline{65.50}} & \multicolumn{1}{c|}{53.12}          & 49.26          \\ \hline

RSG-SCL-LDAM   (ours)                            & \multicolumn{1}{c|}{\textbf{91.28}} & \multicolumn{1}{c|}{84.28}          & 80.50          & \multicolumn{1}{c|}{\textbf{66.2}} & \multicolumn{1}{c|}{\underline{54.74}}          & \underline{50.04}
          \\ \hline

\end{tabular}
\caption{Top-1 accuracy of ResNet-32 on CIFAR-10-LT and CIFAR-100-LT, with the best outcomes highlighted in bold. Results underlined indicate the second-best performance. $\dagger$ indicates results from \cite{Hybrid_contrastive}.}
\label{tab:CIFAR}
\end{table*}

\noindent In our experiments, we observed that the model exhibits insensitivity to the weights \(\eta\) for \(\mathcal{L}_{CESC}\) and \(\mu\) for \(\mathcal{L}_{MV}\). The \cite{RSG} had initially set these weights at \(\eta=0.1\) and \(\mu=0.01\). However, following our scaling these two weights, which aimed to investigate the impact of these parameters, we found that adjustments in \(\eta\) and \(\mu\) did not significantly influence the model's results. Therefore, to ensure a balanced contribution of each loss function to the total loss and maintain the loss functions within the same order of magnitude, we set \(\eta=0.00001\) and \(\mu=0.000001\) for all our experiments. The hyperparameter \(\tau\) for the \(\mathcal{L}_{SCL}\) loss is set to 0.1, aligning with the configurations in \cite{Balanced_contrastive}. For the \(\mathcal{L}_{LDAM}\) loss, the maximum margin (\(max\_m\)) and scaling factor (\(s\)) are set to 0.5 and 30, respectively, following the parameter settings suggested in \cite{LDAM,RSG,survey}. All our experiments were conducted on a NVIDIA GeForce RTX 4090. The source code will be made available upon acceptance.

\subsection{Datasets}
\noindent In our experiments, we simulate long-tailed distributions by down-sampling the original datasets exponentially, guided by imbalance factors $\beta$ defined by $\beta=N_{max}/N_{min}$, where $N_{max}$ and $N_{min}$ represent the largest and smallest class sizes, respectively.

\textbf{Long-Tailed CIFAR-10 and CIFAR-100} CIFAR-10-LT and CIFAR-100-LT are subsets sampled from CIFAR-10 and CIFAR-100 \cite{CIFAR}, respectively. Both CIFAR-10 and CIFAR-100 contain 50,000 images for training and 10,000 images for validation, with 10 and 100 classes respectively. Every image has a size of $32\times32$ pixels. For a fair comparison, we utilize long-tail versions of datasets in line with \cite{Hybrid_contrastive,Balanced_contrastive,BBN,Good_and_Bad}, with imbalance factors $\beta$ set to values of 10, 50, and 100.

\textbf{Long-Tailed mini-ImageNet} The mini-ImageNet \cite{mini-ImageNet} dataset, a subset of the ImageNet \cite{ImageNet}, consists of 100 classes with 600 images each, where every image measures 84×84 pixels. We allocate 48,000 images for the training set, 6,000 for validation, and another 6,000 for testing. We resample the mini-ImageNet dataset according to the imbalance factor $\beta=50$ to achieve a long-tailed distribution, named mini-ImageNet-LT.

\textbf{ImageNet-LT} ImageNet-LT, proposed by \cite{ImageNet-LT}, represents a skewed version of the standard ImageNet dataset \cite{ImageNet}, constructed by adopting a subset that adheres to a Pareto distribution characterized by a power value of $\alpha = 0.6$. This dataset encompasses a diverse collection of 115.8K images, spanning 1000 distinct classes. The number of images per class varies significantly, ranging from a maximum of 1280 to a minimum of just 5 images, thereby creating a long-tailed distribution of samples across classes.

\begin{table*}[!h]
\centering
\begin{tabular}{l|c|c|c|c|c|c|c|c|c|c|c}
\hline
\bf{Label}           & \bf{0}             & \bf{1}             & \bf{2}             & \bf{3}             & \bf{4}             & \bf{5}             & \bf{6}             & \bf{7}             & \bf{8}             & \bf{9}             & \bf{AVG.}           \\ \hline
Sample Size     & 5000          & 2997          & 1796          & 1077          & 645           & 387           & 232           & 139           & 83            & 50            &               \\ \hline\hline
ICD without RSG & \textbf{1.45} & \textbf{2.63} & 1.47          & 1.52          & 1.25          & 1.58          & 1.12          & 1.52          & 1.24          & 1.73          & 1.55          \\ \hline
ICD with RSG    & 1.50          & 2.72          & \textbf{1.40} & \textbf{1.45} & \textbf{1.05} & \textbf{1.39} & \textbf{0.97} & \textbf{1.08} & \textbf{1.02} & \textbf{1.32} & \textbf{1.39} \\ \hline
\end{tabular}
\caption{The impact of RSG on ten classes of Intra-class distance (ICD) in CIFAR-10-LT with an imbalance factor of 100. AVG denotes the average ICD across all classes.}
\label{tab:ICD}
\end{table*}

\subsection{Implementation details}
\noindent For both CIFAR-10-LT and CIFAR-100-LT datasets, we used the ResNet-32 \cite{resnet} as the backbone. We trained the model for 200 epochs with a batch size of 32. The epoch threshold $T_{th}$ of RSG is 100. We utilized the SGD optimizer with a momentum of 0.9 and a weight decay of 5e-4. Our learning rate decay strategy is referenced from \cite{Balanced_contrastive}. The learning rate was initially set to warm up to 0.1 within the first 5 epochs. After the warm-up period, a cosine annealing schedule was used to gradually reduce the learning rate. The learning rate also underwent a stepwise decay at epochs 160 and 180, with each step reducing the learning rate by a factor of 0.1. We do not use class re-weighting strategies, as using inverse class frequency to weight $\mathcal{L}_{LDAM}$ impaired the model's performance. Through genetic algorithms \cite{genetic_1,genetic_2}, we have determined relatively good weights $\alpha$ and $\lambda$ for $\mathcal{L}_{SCL}$ and $\mathcal{L}_{LDAM}$, respectively, with specific values presented in Table \ref{tab:params_for_loss_weights}.

\begin{table}[!h]
\centering
\begin{tabular}{c|ccc|ccc}
\hline
              \bf{Param.}   & \multicolumn{3}{c|}{\bf{CIFAR-10-LT}}                                & \multicolumn{3}{c}{\bf{CIFAR-100-LT}}                               \\ \hline
$\mathbf{\beta}$ & \multicolumn{1}{c|}{\bf 10}    & \multicolumn{1}{c|}{\bf 50}    & \bf 100   & \multicolumn{1}{c|}{\bf 10}    & \multicolumn{1}{c|}{\bf 50}    & \bf{100}   \\ \hline
$\alpha$         & \multicolumn{1}{c|}{1.969} & \multicolumn{1}{c|}{9.764} & 6.299 & \multicolumn{1}{c|}{8.189} & \multicolumn{1}{c|}{8.819} & 8.976 \\ \hline
$\lambda$        &  \multicolumn{1}{c|}{0.079} & \multicolumn{1}{c|}{2.520} & 0.709 & \multicolumn{1}{c|}{0.787} & \multicolumn{1}{c|}{0.315} & 0.472 \\ \hline 
\end{tabular}
\caption{Different weight parameters ($\alpha$ and $\lambda$) representing the well-tuned weights for the $\mathcal{L}_{SCL}$ and the $\mathcal{L}_{LDAM}$, respectively, under the imbalance factor ($\beta$).}
\label{tab:params_for_loss_weights}
\end{table}

For the mini-ImageNet-LT dataset, we utilize a pre-trained ResNeXt-50-32x4d \cite{resnext} model as the backbone. The model is fine-tuned over 20 epochs with a batch size of 32. The epoch threshold $T_{th}$ of RSG is 10. We adopt the SGD optimizer with an initial learning rate of 0.001, momentum of 0.9, and a weight decay of 5e-4. A StepLR scheduler is applied to adjust the learning rate with a decay factor of 0.9 after each epoch. The weight of $\mathcal{L}_{SCL}$, $\alpha$, is set to 0.9, and that of $\mathcal{L}_{LDAM}$, $\lambda$, to 0.1. Additionally, the weight assigned to Cross-Entropy (CE) loss is consistent with that of LDAM.

For ImageNet-LT dataset, we employed the ResNeXt-50-32x4d model as the backbone. We trained the model for 90 epochs with a batch size of 128. The epoch threshold $T_{th}$ of RSG is 45. The learning rate was initially set to warm up to 0.03 within the first 5 epochs. After the warm-up period, a cosine annealing schedule was used to reduce the learning rate gradually. We utilized the SGD optimizer with a momentum of 0.9 and a weight decay of 5e-4. Regarding the loss function weights, we set $\alpha$ at 0.35 and $\lambda$ at a weight of 1.


To assess the model's performance on classes with different sample sizes, we divided the classes into three groups based on the number of samples: Many-shot (more than 100 images), Medium-shot (20 to 100 images), and Few-shot (less than 20 images). Inspired by paper \cite{genetic_1,genetic_2}, we used genetic algorithms to automatically find relatively optimal values for part of the model's hyperparameters.

\subsection{Ablation study}
\noindent In our experiments, we distinguish between two configurations: SCL-LDAM, where SCL is used for feature learning and LDAM for classification without RSG integration, and RSG-SCL-LDAM, where we integrate RSG into the model with SCL for feature learning and LDAM for classification. Previous findings \cite{survey,RSG} indicate that RSG aids in improving tail class accuracy when combined with CE and LDAM, often at the cost of head class accuracy. However, our analysis in Section \ref{sec:analysis_and_methods} posits that RSG can collaborate with SCL and LDAM, offsetting their respective limitations. This ablation study aims to confirm if RSG can enhance tail class precision without detriment to head class accuracy, thus proving its collaborative efficacy with SCL and LDAM.

Besides, on the mini-ImageNet dataset, we conducted tests ablationing SCL from RSG-SCL-LDAM, the RSG-LDAM, to evaluate the impact of SCL on the model. We also experimented with replacing LDAM with Cross-Entropy (CE) loss, the RSG-SCL-CE, to observe the individual contribution of LDAM.

\subsection{Main results}
\noindent\textbf{Experimental results on long-tailed CIFAR} Based on the experimental results in Table \ref{tab:CIFAR}, the RSG-SCL-LDAM outperforms most others. Compared to Hybrid-SC, which uses CE for classification and SCL for representation learning, SCL-LDAM employs the LDAM classifier, which is more sensitive to tail classes. However, it falls short of Hybrid-SC in scenarios with imbalance factors of 50 and 100 on CIFAR-10-LT, which can be attributed to not employing the most ideal loss weighting between SCL and LDAM. Although most of our results did not exceed BCL utilizing similiar framework with ours, it's noteworthy that it utilizes data augmentation techniques that expand the dataset with additional images. Specifically, the three different views generated from each image through data augmentation contribute high accuracy to BCL also tripled the computational cost, as shown in Table \ref{tab:params_FLOPs}. The SCL shows strong performance in classes with abundant samples, and it is consistently allocated a higher weight in our method, indicated in Table \ref{tab:params_for_loss_weights}. Thus, SCL contributes to our method achieving the highest accuracy on datasets with an imbalance factor of 10.

\begin{table}[!h]
\centering
\begin{tabular}{c|c|c|c}
\hline
           & BCL\cite{Balanced_contrastive}                                                & SCL-LDAM  
           & RSG-SCL-LDAM                                                 \\ \hline
\# params. & $2.17\times10^7$ & $2.01\times 10^7  $ & $2.08 \times 10^7  $\\ \hline
FLOPs      & $3.31 \times 10^{11}$& $1.10\times 10^{11}$ & $1.12 \times 10^{11}$ \\ \hline
\end{tabular}
\caption{The numbers of parameters and
FLOPs of the BCL and our method.}
\label{tab:params_FLOPs}
\end{table}

Furthermore, results shown in Table \ref{tab:CIFAR100-IF100} demonstrate that our method still performs well. Notably, previous studies \cite{RSG,survey} integrating RSG into models to enhance Few-shot class accuracy often compromised the accuracy of Many-shot classes. In contrast, our results indicate an overall improvement. We introduced the Intra-class distance (ICD), which measures how tightly grouped the samples are around their class center in the feature space. Specifically, computing the ICD begins by determining the feature center for each class, which involves averaging the feature vectors of all samples within the same class. Then, for each sample, the Euclidean distance between its feature vector and the feature center of its class is calculated. This distance measures how far the sample deviates from the center in the feature space. Finally, the class ICD is obtained by averaging these distances across all samples. We computed the ICD for the CIFAR-10-LT dataset with an imbalance factor of 100 in Table \ref{tab:ICD}. With the exception of the predetermined frequent classes labeled 0 and 1, all other classes became more clustered due to the introduction of RSG, without significantly affecting the cohesiveness within classes 0 and 1. Hence, we believe that expanding the feature space for tail classes does not necessarily impair the accuracy of head classes when there is intra-class clustering and clear inter-class separation.

\begin{table}[h]
\centering
\begin{tabular}{l|c|c|c|c}
\hline
\bf{Method}                  & \bf{Many}          & \bf{Medium}        & \bf{Few}           & \bf{All}           \\ \hline
$\tau$-norm$^\dagger$\cite{tau-norm} & 61.4          & 42.5          & 15.7          & 41.4          \\ \hline
Hybrid-SC\cite{Hybrid_contrastive}                & -             & -             & \textbf{-}    & 46.7          \\ \hline
MetaSAug-LDAM           & \textbf{-}    & \textbf{-}    & -             & 48            \\  
\cite{Metasaug}  & & & &  \\ \hline
DRO-LT\cite{DROLT}                   & 64.7          & 50            & 23.8          & 47.3          \\ \hline
RIDE (3 experts)\cite{RIDE}          & \textbf{68.1} & 49.2          & 23.9          & 48            \\ \hline
BCL\cite{Balanced_contrastive}                      & \underline{67.2}          & \textbf{53.1} & \textbf{32.9} & \textbf{51.9} \\ \hline\hline
SCL-LDAM                         & 66.6          & 51.7         & 27.1         & 49.3        \\ \hline
RSG-SCL-LDAM                           & 66.9          & \underline{52.2}        & \underline{28.6}         & \underline{50.1}        \\ \hline
\end{tabular}
\caption{Top-1 accuracy of ResNet-32 on CIFAR-100-LT with an imbalance factor of 100 running for 200 epochs. Results that are bolded indicate the best performance, while those underlined represent the second-best performance. $\dagger$ indicates results from \cite{DROLT}.}
\label{tab:CIFAR100-IF100}
\end{table}

\noindent\textbf{Experimental results on long-tailed mini-ImageNet} The results in Table \ref{tab:mini-ImageNet} indicate that RSG-SCL-LDAM comprehensively improve accuracy across all classes. Ablating RSG leads to a decline in model precision for every class, underscoring the beneficial impact of integrating RSG on the overall classification performance of the model. Furthermore, leveraging the SCL branch for representation learning leads to performance gains, while replacing CE with LDAM specifically increases accuracy significantly for few classes and also shows improvements for the rest.

\begin{table}[h]
\centering
\begin{tabular}{l|c|c|c|c}
\hline
\multicolumn{1}{l|}{\bf Methods} & \bf Many           & \bf Medium         & \bf Few            & \bf{All}            \\ \hline
RSG-LDAM                      & 94.56          & 88.43          & 75.42          & 92.75          \\ \hline
RSG-SCL-CE                    & 96.01          & 90.59          & 77.08          & 94.45          \\ \hline\hline
SCL-LDAM                     & 95.80          & 94.21          & 93.17          & 95.22          \\ \hline
RSG-SCL-LDAM                 & \textbf{96.03} & \textbf{94.80} & \textbf{95.42} & \textbf{95.61} \\ \hline

\end{tabular}
\caption{Top-1 accuracy of pre-trained ResNeXt-50-32x4d on mini-ImageNet-LT with an imbalance factor of 50 running for 20 epochs. }
\label{tab:mini-ImageNet}
\end{table}

\begin{table}[h]
\centering
\begin{tabular}{l|c|c|c|c}
\hline
\bf Methods                  & \bf Many & \bf Medium & \bf Few  & \bf All  \\ \hline
Softmax$^\dagger$               & 66.5 & 39.0   & 8.6  & 45.5 \\ \hline
Focal Loss\cite{Focal}               & 64.3 & 37.1   & 8.2  & 43.7 \\ \hline
LDAM\cite{LDAM}               & 65.6 & 43.1   & 18.9  & 48.5 \\ \hline
$\tau$-norm\cite{tau-norm} & 59.1 & 46.9   & 30.7 & 49.4 \\ \hline

LWS\cite{tau-norm}                      & 60.2 & 47.2   & 30.3 & 49.9 \\ \hline
LADE\cite{LADE}                     & 62.3 & 49.3   & 31.2 & 51.9 \\ \hline
Causal model\cite{Good_and_Bad}             & 62.7 & 48.8   & \underline{31.6} & 53.4 \\ \hline
DisAlign\cite{DisAlign}                 & 62.7 & \underline{52.1}   & 31.4 & 53.4 \\ \hline
RIDE (2 experts)\cite{RIDE}          & -    & -      & -    & \underline{55.9} \\ \hline
BCL\cite{Balanced_contrastive}                      & \textbf{67.2} & \textbf{53.9}   & \textbf{36.5} & \textbf{56.7} \\ \hline\hline
SCL-LDAM (ours)     &    \underline{66.8}        &   44.5   &   21.9     &     50.1      \\ \hline
RSG-SCL-LDAM (ours)   &     66.6        &   45.7   &     23.2   &      50.5      \\ \hline
\end{tabular}
\caption{Top-1 accuracy of ResNeXt-50-32x4d on ImageNet-LT running for 90 epochs. Results that are bolded indicate the best performance, while those underlined represent the second-best performance. $\dagger$ indicates results from \cite{survey}.}
\label{tab:ImageNet-LT}
\end{table}

\noindent \textbf{Experimental results on ImageNet-LT} According to the results in Table \ref{tab:ImageNet-LT}, most methods improve the accuracy of Medium-shot and Few-shot classes at the expense of Many-shot class accuracy compared to the baseline Softmax established in the paper \cite{survey}. In contrast, our method achieves a comprehensive enhancement across all classes. Compared to two-stage training methods such as LWS \cite{tau-norm}, $\tau$-norm \cite{tau-norm}, and DisAlign \cite{DisAlign}, and ensemble learning approaches like RIDE \cite{RIDE}, our approach is notably simpler, almost not adding extra computational complexity and parameters to the model. On the ImageNet-LT dataset, which encompasses 1000 classes, Many-shot includes 391 classes, Medium-shot 463, and Few-shot 146. Many-shot classes constitute over one-third of the dataset and have more samples. Given the importance of Many-shot classes, we believe it is crucial to explore methods that balance accuracy improvements across all classes. BCL \cite{Balanced_contrastive} is the combination of improved Supervised Contrastive Learning with logits adjustment methods. Despite its increased computational complexity, as indicated in Table \ref{tab:params_for_loss_weights}, BCL effectively enhances accuracy across all classes. Therefore, the result of BCL further validates that collaboratively using SCL and logits adjustment is an effective strategy for addressing long-tail recognition challenges.

The result for LDAM in Table \ref{tab:ImageNet-LT} was obtained from our experiments, designed to observe the enhancements by SCL and RSG. Notably, we did not employ loss re-weighting, deferred re-weighting, or deferred re-sampling \cite{LDAM,RSG} for LDAM. Comparing with LDAM outcomes, we further confirmed that introducing SCL compensates for the accuracy reduction of Many-shot classes caused by LDAM. Due to the larger number of classes and greater imbalance in the ImageNet-LT dataset, achieving intra-class clustering is more challenging. Consequently, integrating RSG reduced the accuracy of Many-shot classes by 0.2\%, but it still outperformed the baseline Softmax.

We applied three data augmentation methods in our experiments: random size cropping, random horizontal flipping, and color jittering. Attempting to reduce the difference between training and validation accuracy, we added additional augmentation techniques but observed decreased accuracy. We believe excessive data augmentation leads to further dispersion of features within each class, which is particularly detrimental for originally dispersed tail classes, contradicting our goal of achieving tighter intra-class feature clustering. Hence, selecting suitable data augmentation methods is important for models based on Supervised Contrastive Learning, especially without increasing the dataset size. Moreover, during training, we tested three learning rate decay strategies: Step Decay, Cosine Annealing, and Cosine Annealing with Warm Restarts. Cosine Annealing achieved the best results.

More information on hyperparameter and learning rate adjustments can be found in Supplementary material.
\section{Conclusion}
\label{sec:conclusion}
\noindent In this work, we explored the synergy and compensatory effects of Supervised Contrastive Learning, Rare-Class Sample Generator, and Label-Distribution-Aware Margin Loss on the long-tail recognition challenge. Our in-depth analysis demonstrates that balancing these three techniques leverages their respective strengths to offset their limitations, enhancing tail class accuracy while maintaining dominant class performance, thereby achieving a balanced improvement across all classes. Extensive experiments validate this collaborative approach's effectiveness in addressing the long-tail learning challenge.

\noindent \textbf{Limitation} Although the weighted linear combination of losses offers a straightforward and effective approach to long-tail recognition, considerable time and computational resources were used in searching for optimal loss weight hyperparameters using genetic algorithms. Suboptimal weight selection can limit the synergy between loss functions, leading to less-than-ideal results. Future work should aim to discover a robust and superior weighting method for combined loss functions.
\clearpage
\setcounter{page}{1}
\maketitlesupplementary

\section*{Determining the Weights of the Loss Function on the CIFAR Datasets}

\begin{figure}[htbp]
    \centering
    \includegraphics[width=\columnwidth]{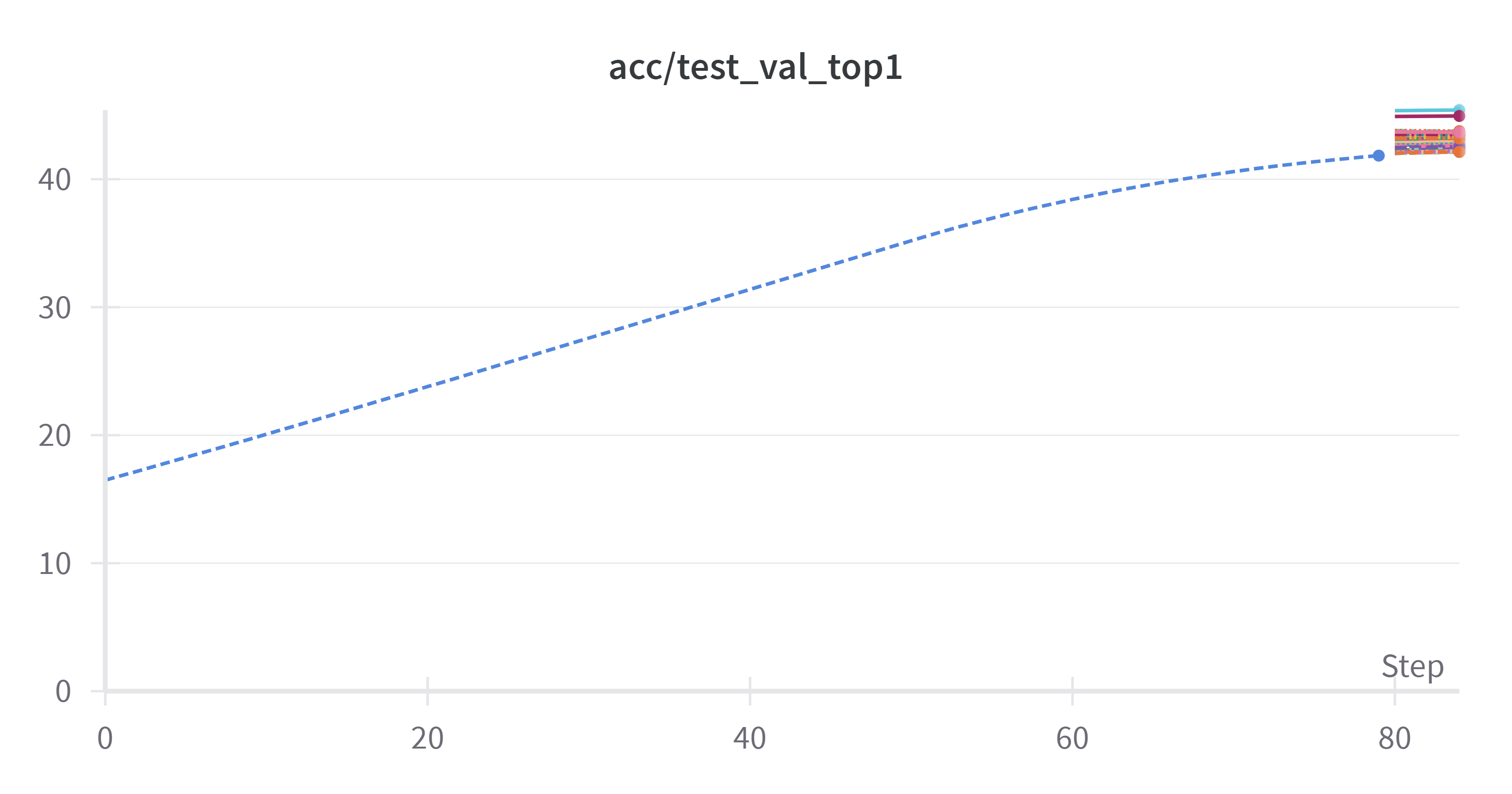}
    \caption{Searching for suitable SCL and LDAM weights for the CIFAR10-LT with an Imbalance Factor of 10. The blue dashed line represents the validation accuracy of our pre-trained model over the first 80 epochs. Epochs 81 to 85 depict the process of searching for hyperparameter weights using the genetic algorithm.}
    \label{fig:cifar10-if10}
\end{figure}

We utilized a genetic algorithm \cite{genetic_1,genetic_2} to find appropriate weights for SCL and LDAM losses that synergize effectively. Due to the extensive time and computational resources required for 200 epochs of experiments, directly applying a genetic algorithm was hard for us. Our approach was initially trained with SCL and LDAM weights set to 1. By monitoring the validation accuracy, we noted that the rate of improvement gradually slowed after 80 epochs. Consequently, we saved the model at this stage to serve as a pre-trained model for our experiments. We then loaded this pre-trained model and applied the genetic algorithm to search within the $[0,10]$ range for suitable weights for SCL and LDAM. We limited training to five epochs for different weight combinations, adopting the combination with the highest validation accuracy as our experiment's weight hyperparameters, as demonstrated in Figure \ref{fig:cifar10-if10}.

\begin{figure}[!htbp]
    \centering
    \includegraphics[width=\columnwidth]{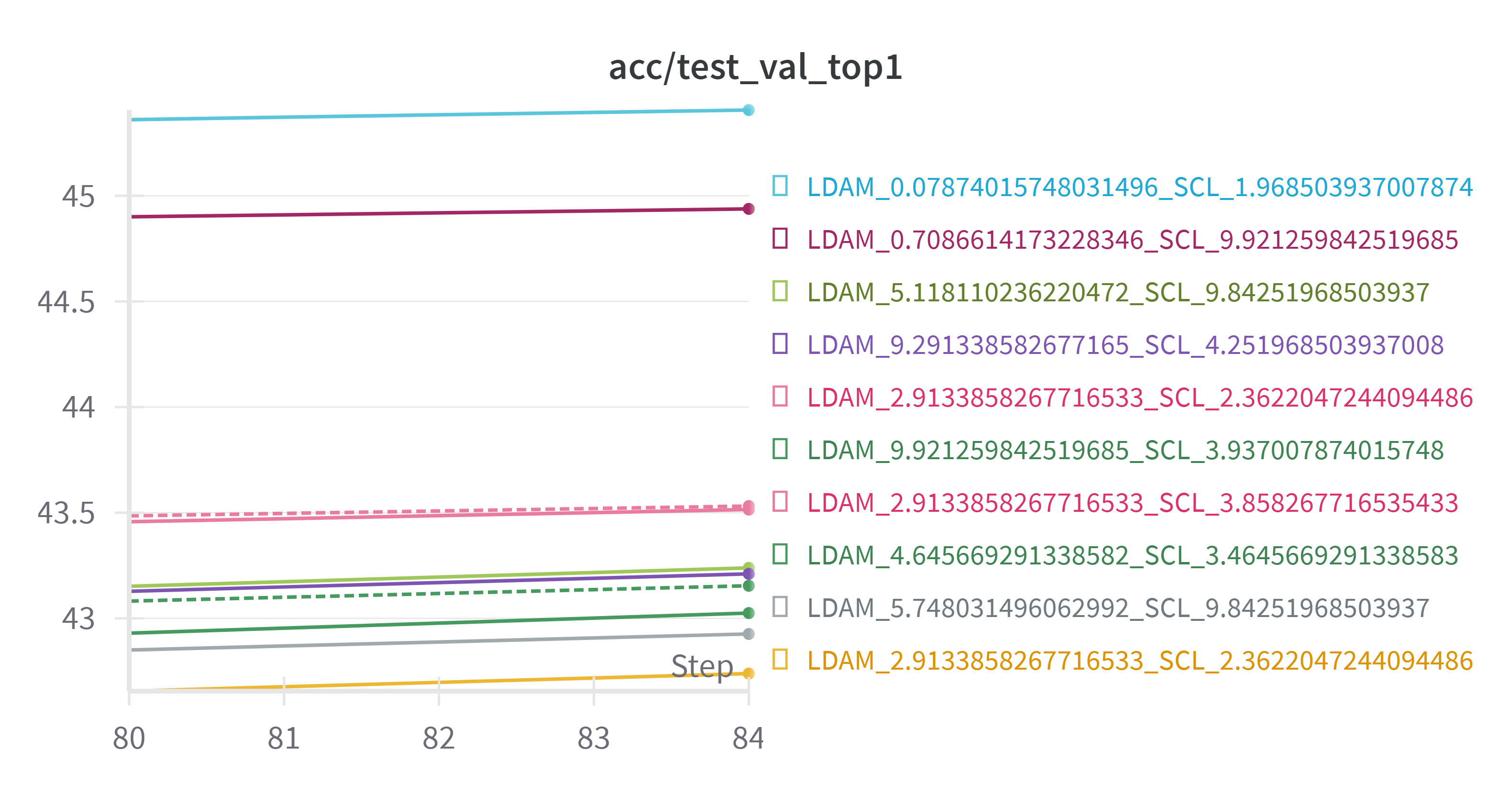}
    \caption{The Top-10 Weight Combinations for Highest Validation Accuracy on the CIFAR10-LT with an Imbalance Factor of 10. In the figure's legend, the naming convention 'LDAM\_X\_SCL\_Y' represents the weight for LDAM loss as X and the weight for SCL loss as Y, where X and Y are specific values obtained from our experiments.}
    \label{fig:cifar10-if10-top10}
\end{figure}

\begin{figure}[!htbp]
    \centering
    \includegraphics[width=\columnwidth]{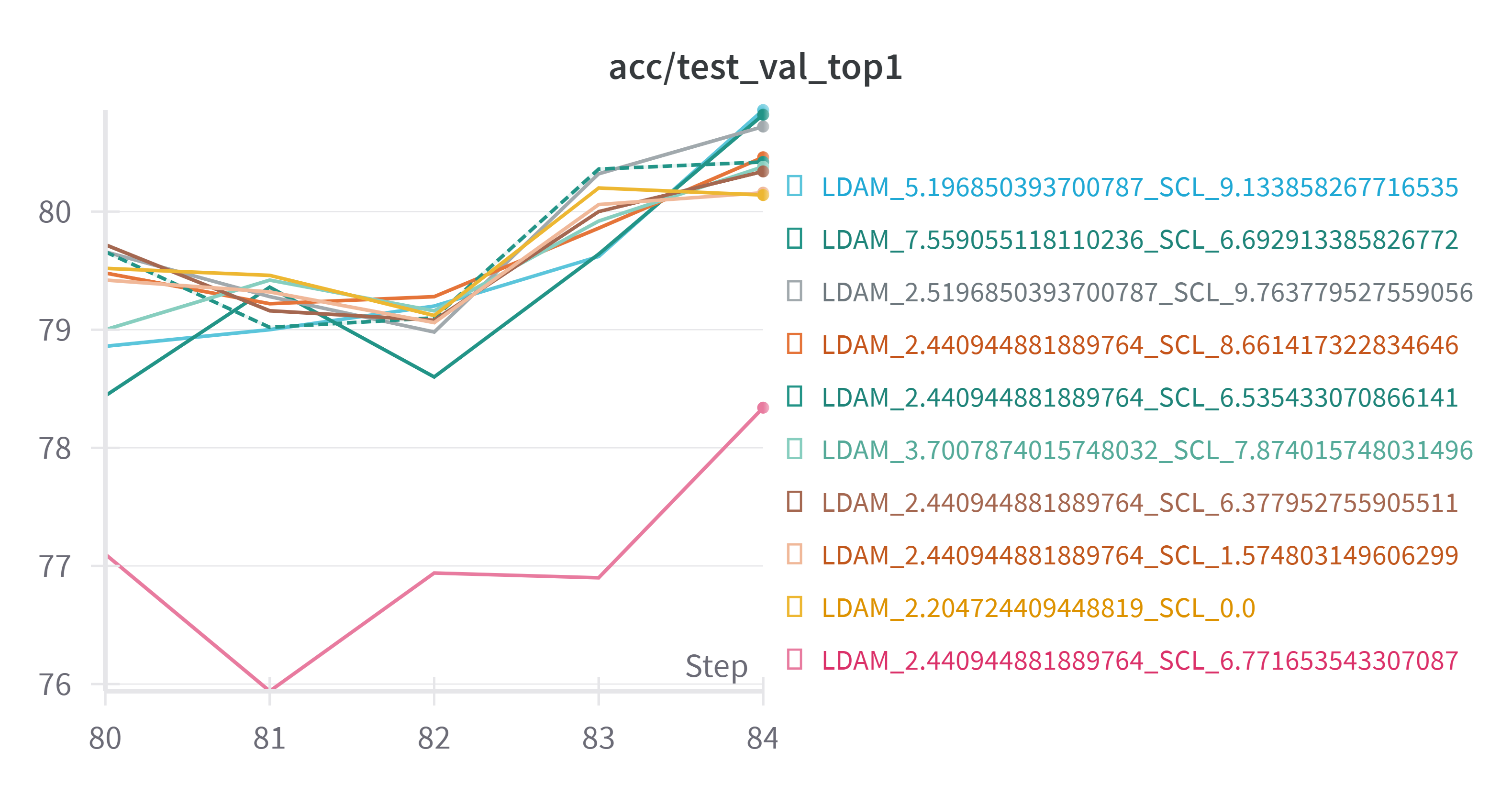}
    \caption{The Top-10 Weight Combinations for Highest Validation Accuracy on the CIFAR10-LT with an Imbalance Factor of 50.}
    \label{fig:cifar10-if50-top10}
\end{figure}

\begin{figure}[!htbp]
    \centering
    \includegraphics[width=\columnwidth]{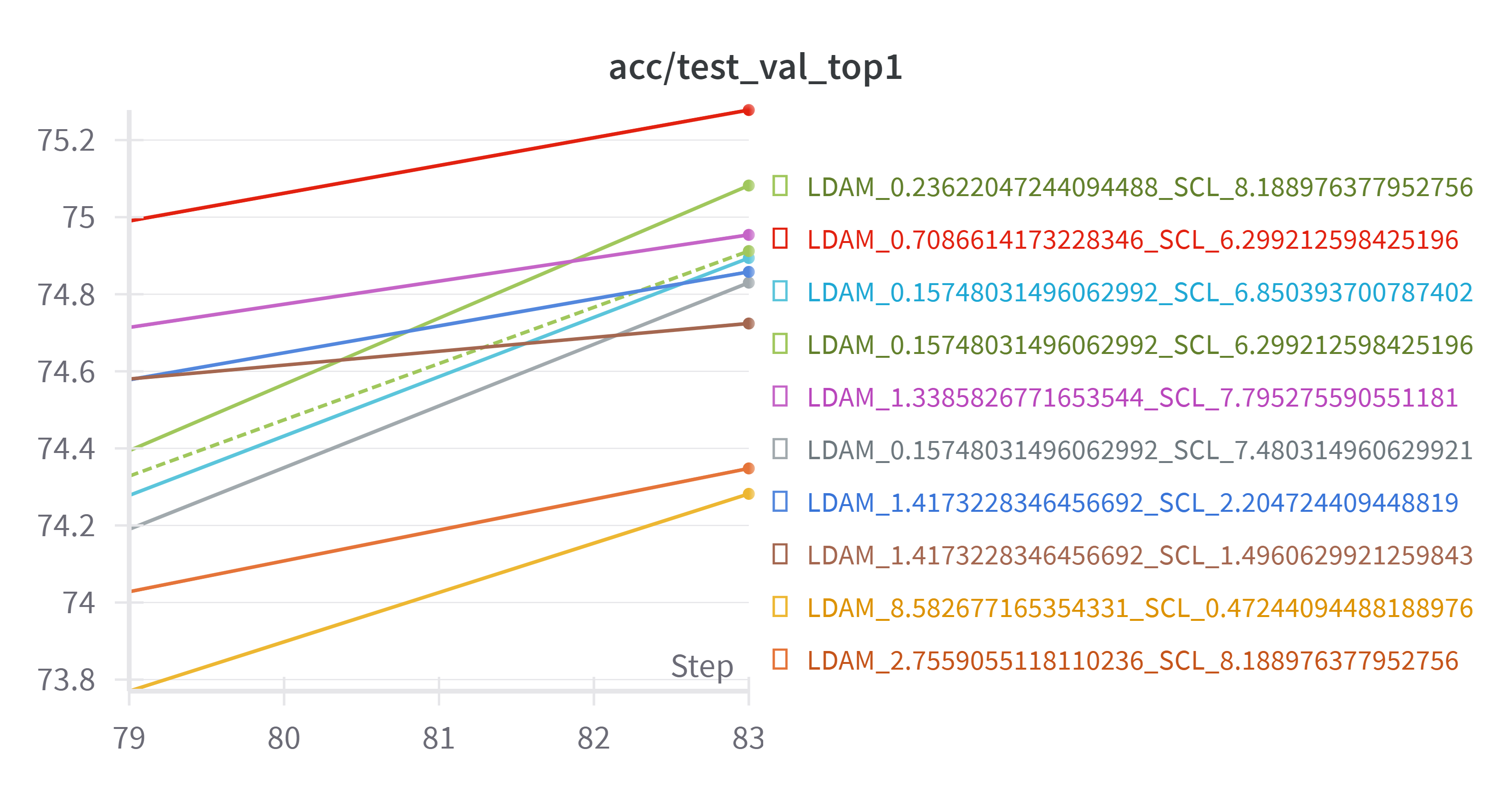}
    \caption{The Top-10 Weight Combinations for Highest Validation Accuracy on the CIFAR10-LT with an Imbalance Factor of 100.}
    \label{fig:cifar10-if100-top10}
\end{figure}

\begin{figure}[!htbp]
    \centering
    \includegraphics[width=\columnwidth]{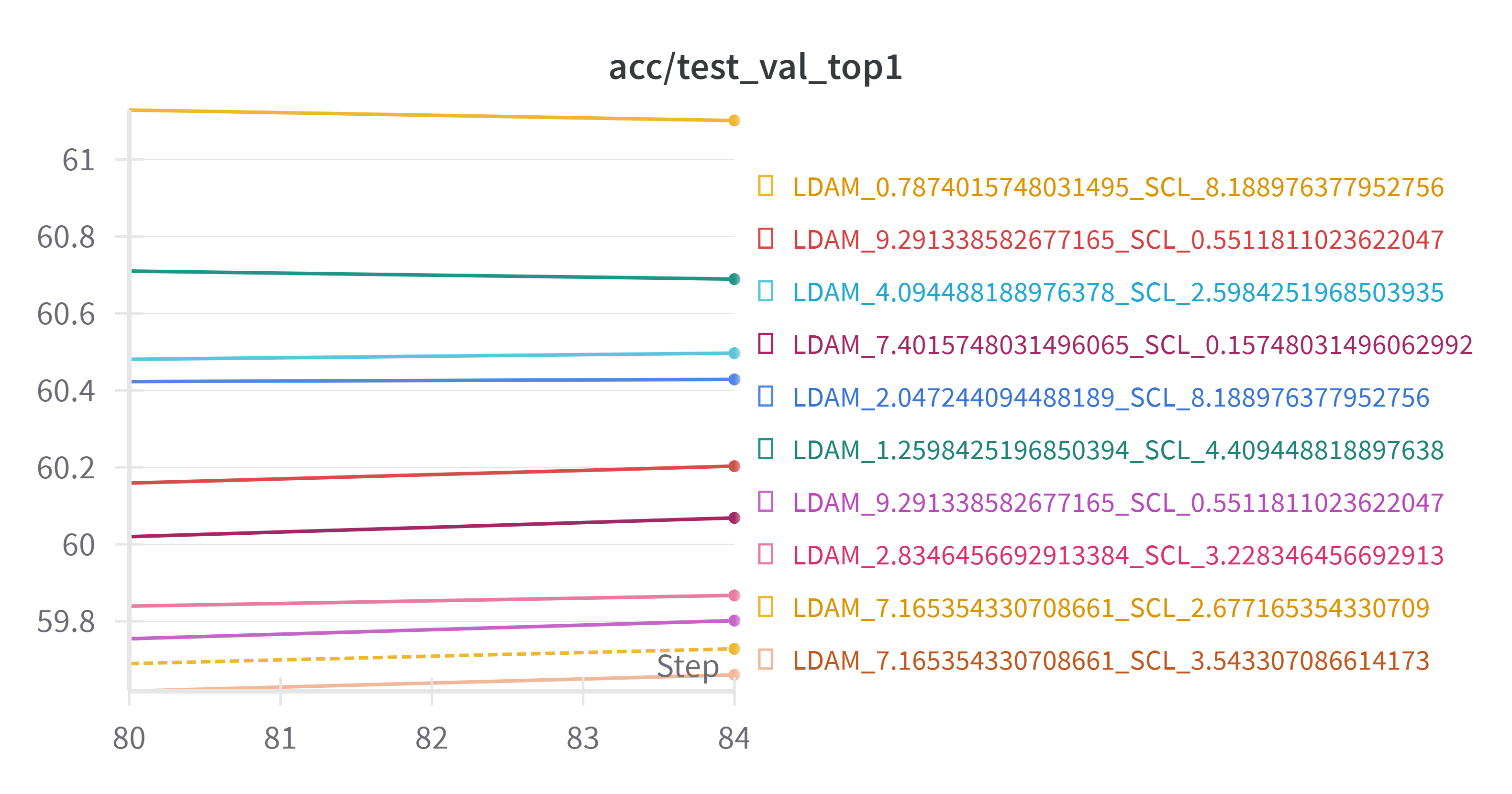}
    \caption{The Top-10 Weight Combinations for Highest Validation Accuracy on the CIFAR100-LT with an Imbalance Factor of 10.}
    \label{fig:cifar100-if10-top10}
\end{figure}

\begin{figure}[!htbp]
    \centering
    \includegraphics[width=\columnwidth]{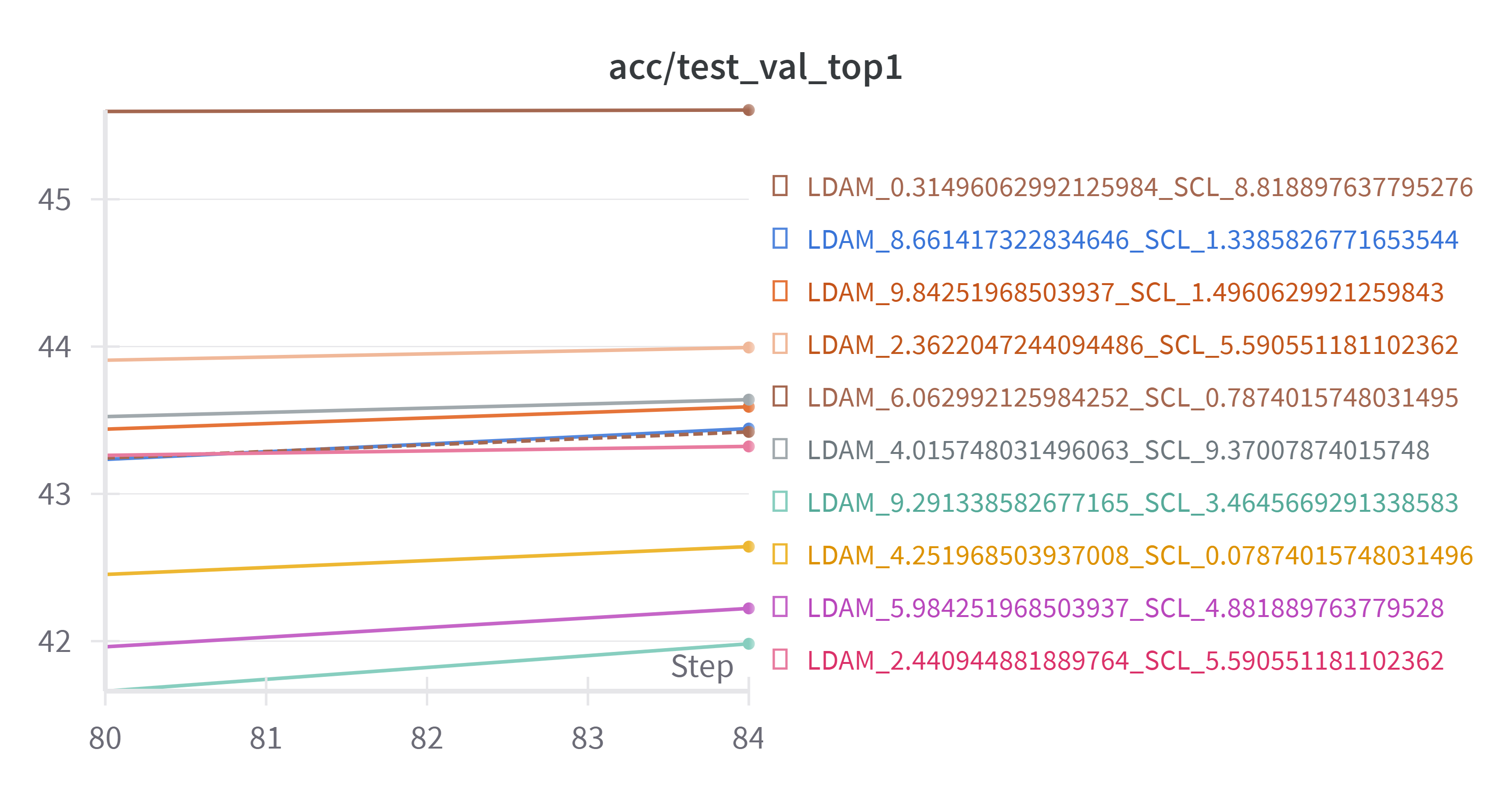}
    \caption{The Top-10 Weight Combinations for Highest Validation Accuracy on the CIFAR100-LT with an Imbalance Factor of 50.}
    \label{fig:cifar100-if50-top10}
\end{figure}

\begin{figure}[!htbp]
    \centering
    \includegraphics[width=\columnwidth]{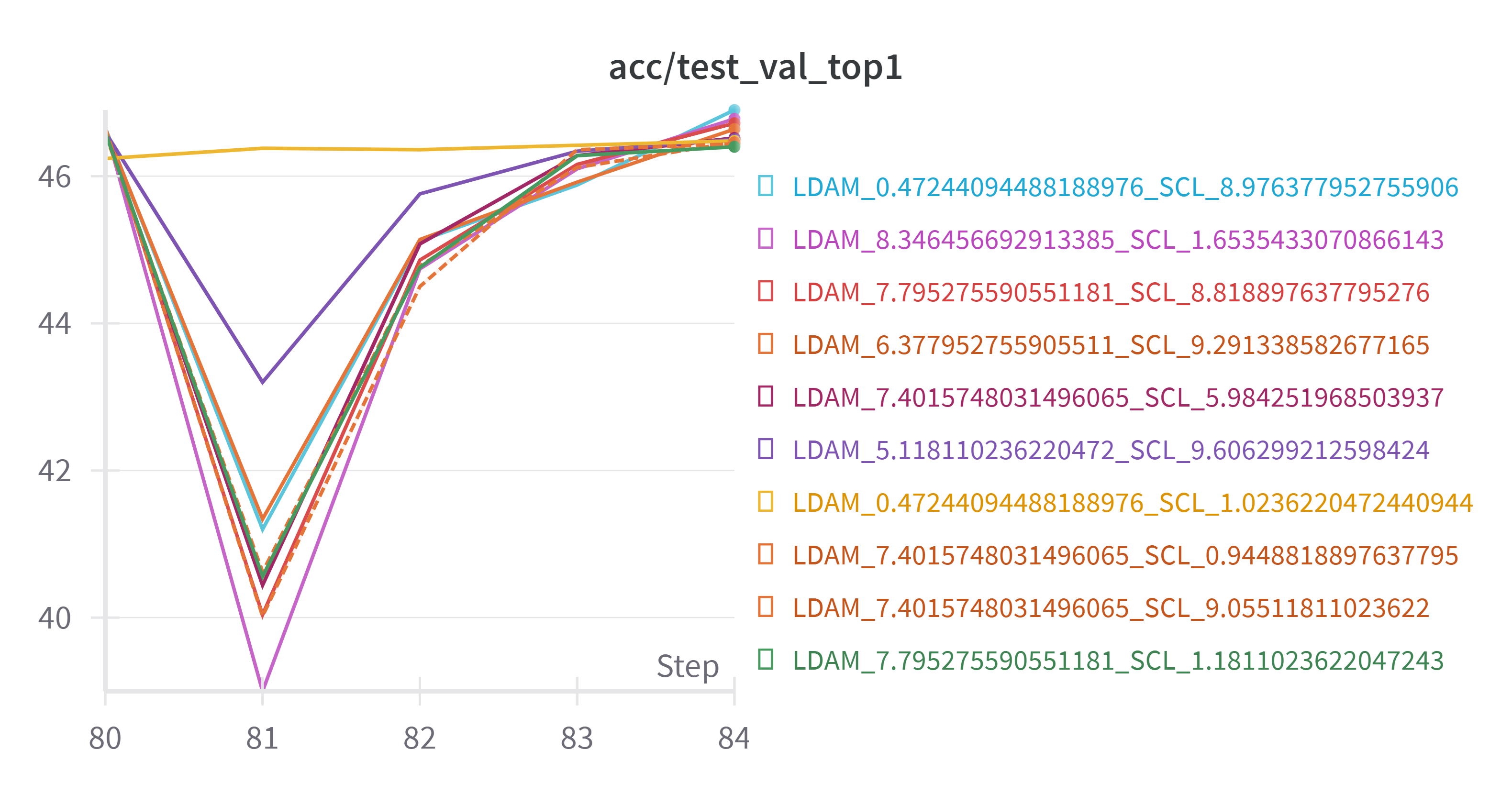}
    \caption{The Top-10 Weight Combinations for Highest Validation Accuracy on the CIFAR100-LT with an Imbalance Factor of 100.}
    \label{fig:cifar100-if100-top10}
\end{figure}

\begin{figure}[!htbp]
    \centering
    \includegraphics[width=\columnwidth]{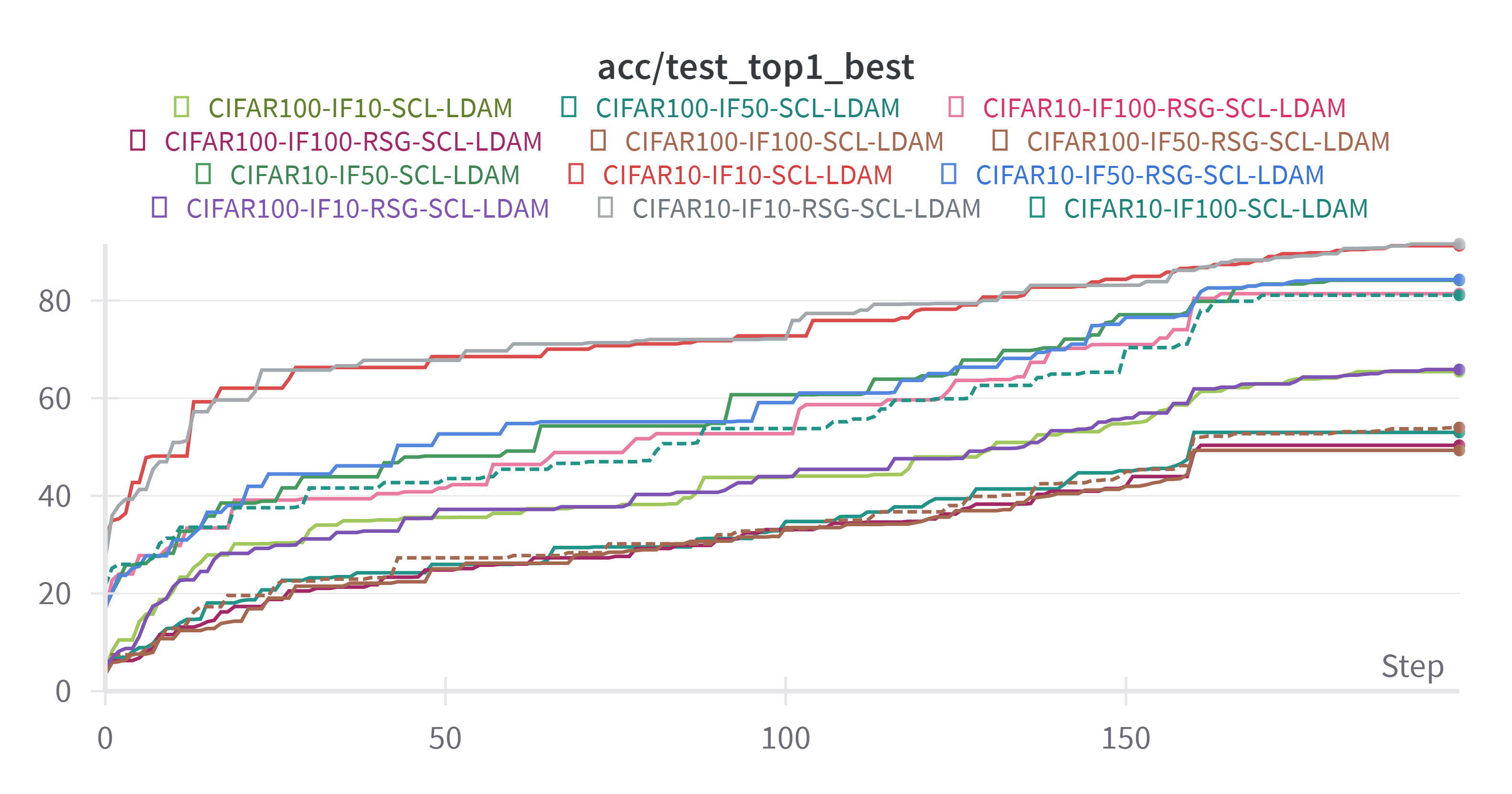}
    \caption{Validation accuracy on the CIFAR10-LT and CIFAR100-LT datasets using the optimal hyperparameter weights.}
    \label{fig:CIFAR}
\end{figure}

We searched for hyperparameter weights in the same manner for three imbalance factors on the CIFAR10-LT and CIFAR100-LT datasets. For clarity, we have omitted the pre-training phase and only display the top ten weight combinations with the highest validation accuracy, as shown in Figures \ref{fig:cifar10-if10-top10} to \ref{fig:cifar100-if100-top10}. The weight combinations yielding the highest accuracy in each figure, rounded to three decimal places, were selected as the experimental hyperparameters. The selected weight parameters are presented in Table \ref{tab:params_for_loss_weights}, and the validation accuracy achieved with these optimal parameters across 200 epochs for both datasets is shown in Figure \ref{fig:CIFAR}.

The yellow line in Figure \ref{fig:cifar10-if50-top10} indicates an SCL weight of 0.0, where the model only optimizes LDAM loss. Yet, in the Top-10 weight combinations, eight surpass this approach. Thus, Figure \ref{fig:cifar10-if50-top10} clearly shows the superiority of the SCL-LDAM combination over LDAM alone for long-tail recognition problems.

Without time constraints, genetic algorithms could lead our hyperparameter to approach the optimal solution. However, a limited timeframe limit us to finding a good solution only, impacting the solution's quality, as discussed in Section \ref{sec:conclusion}. Additionally, we employed pre-trained models and tested each set of hyperparameters for only five epochs. Therefore, the hyperparameters we present also have certain limitations.

\section*{Determining the Learning Rate and Learning Rate Decay Strategy on the ImageNet-LT Dataset}

\begin{figure}[htbp]
    \centering
    \includegraphics[width=\columnwidth]{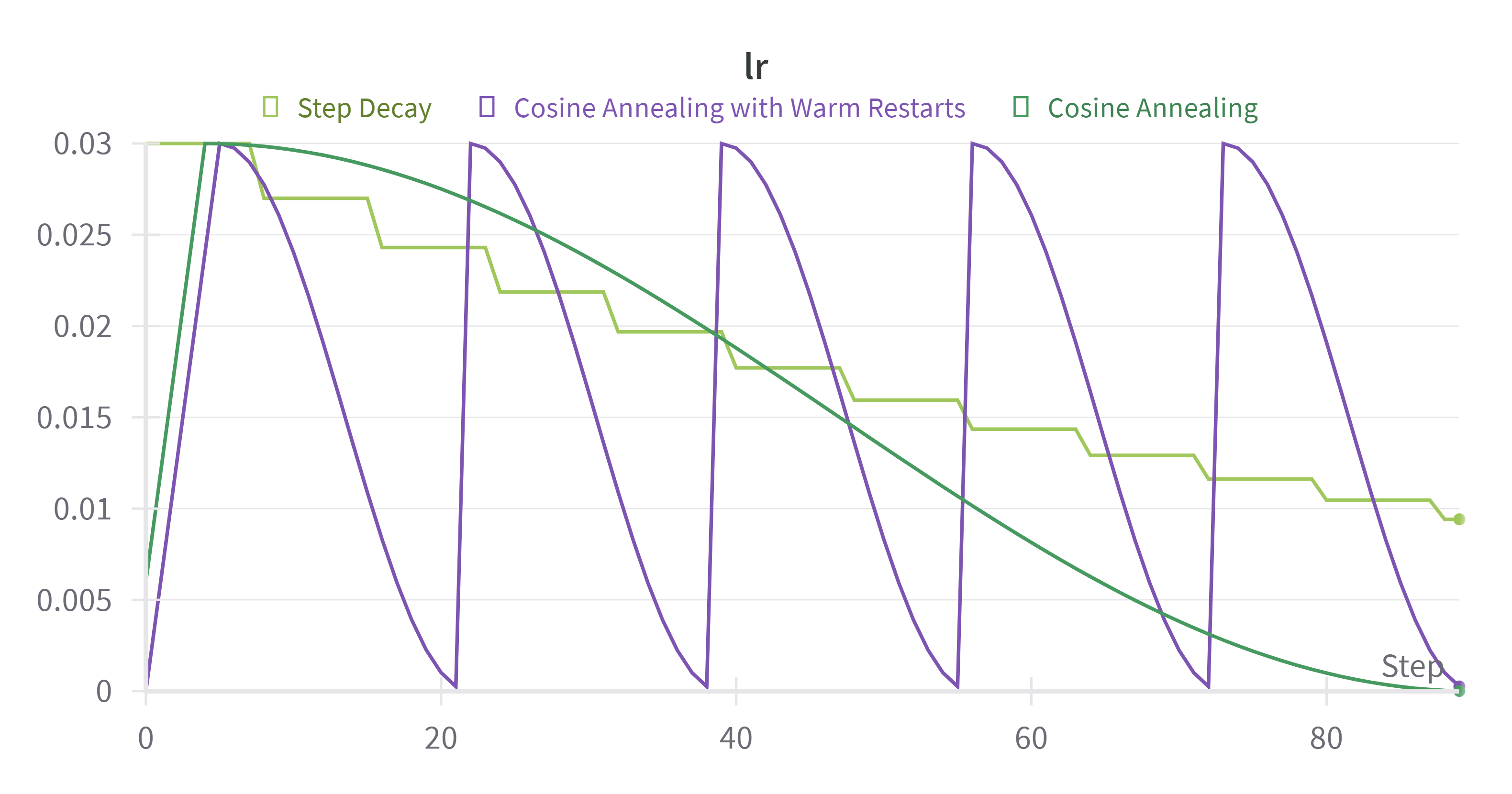}
    \caption{Comparing Different Learning Rate Decay Strategies.}
    \label{fig:I-lr}
\end{figure}

\begin{figure}[htbp]
    \centering
    \includegraphics[width=\columnwidth]{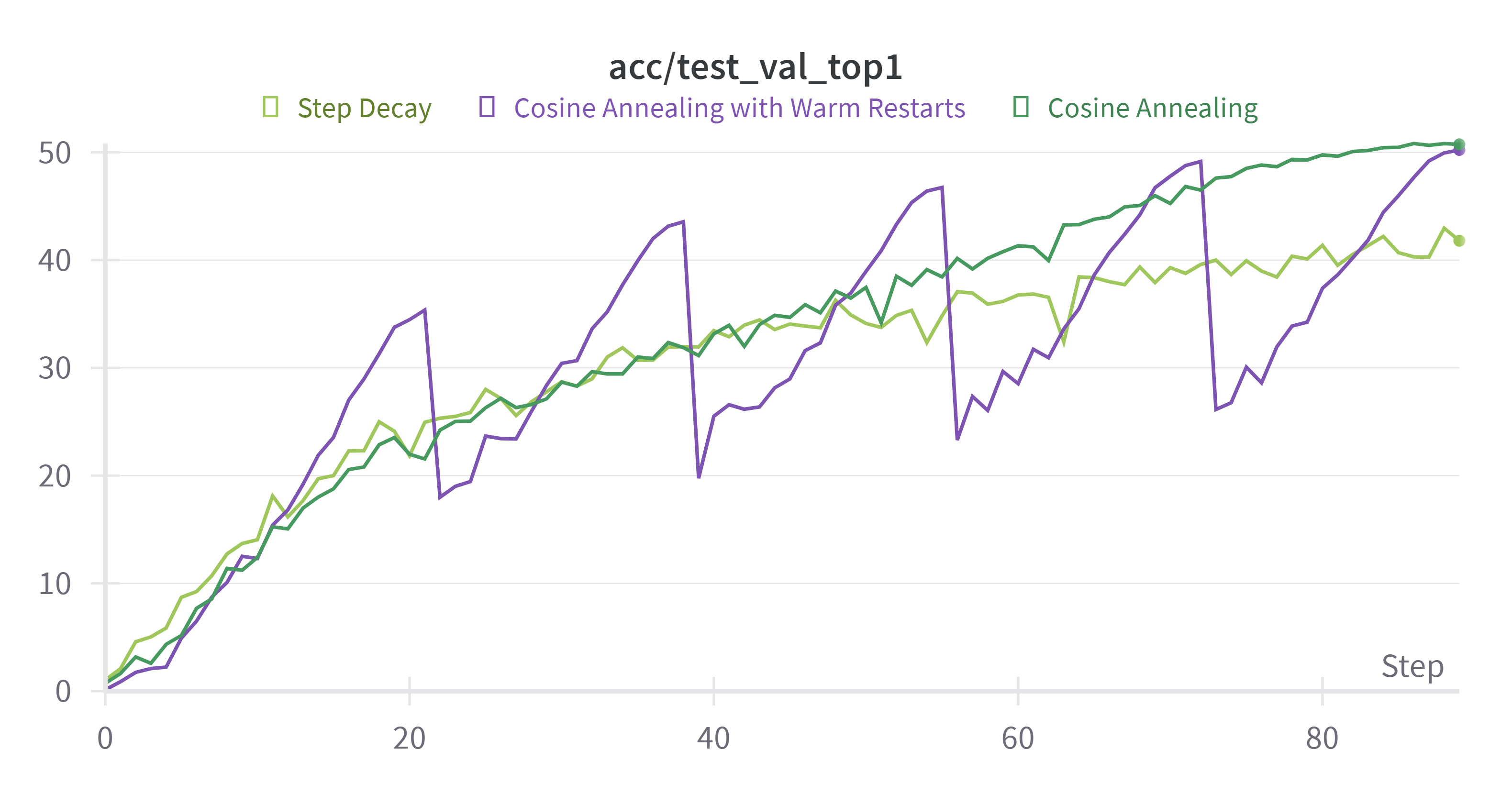}
    \caption{Performance Comparison of Three Learning Rate Decay Strategies on the Validation Set over Training Epochs.}
    \label{fig:I-lr-val-process}
\end{figure}

On the ImageNet-LT dataset, we evaluated three learning rate decay strategies, as illustrated in Figure \ref{fig:I-lr}. Maintaining consistent configurations across all experiments, the results shown in Figure\ref{fig:I-lr-val-process} indicate that Cosine Annealing outperforms other strategies.

\begin{figure}[htbp]
    \centering
    \includegraphics[width=\columnwidth]{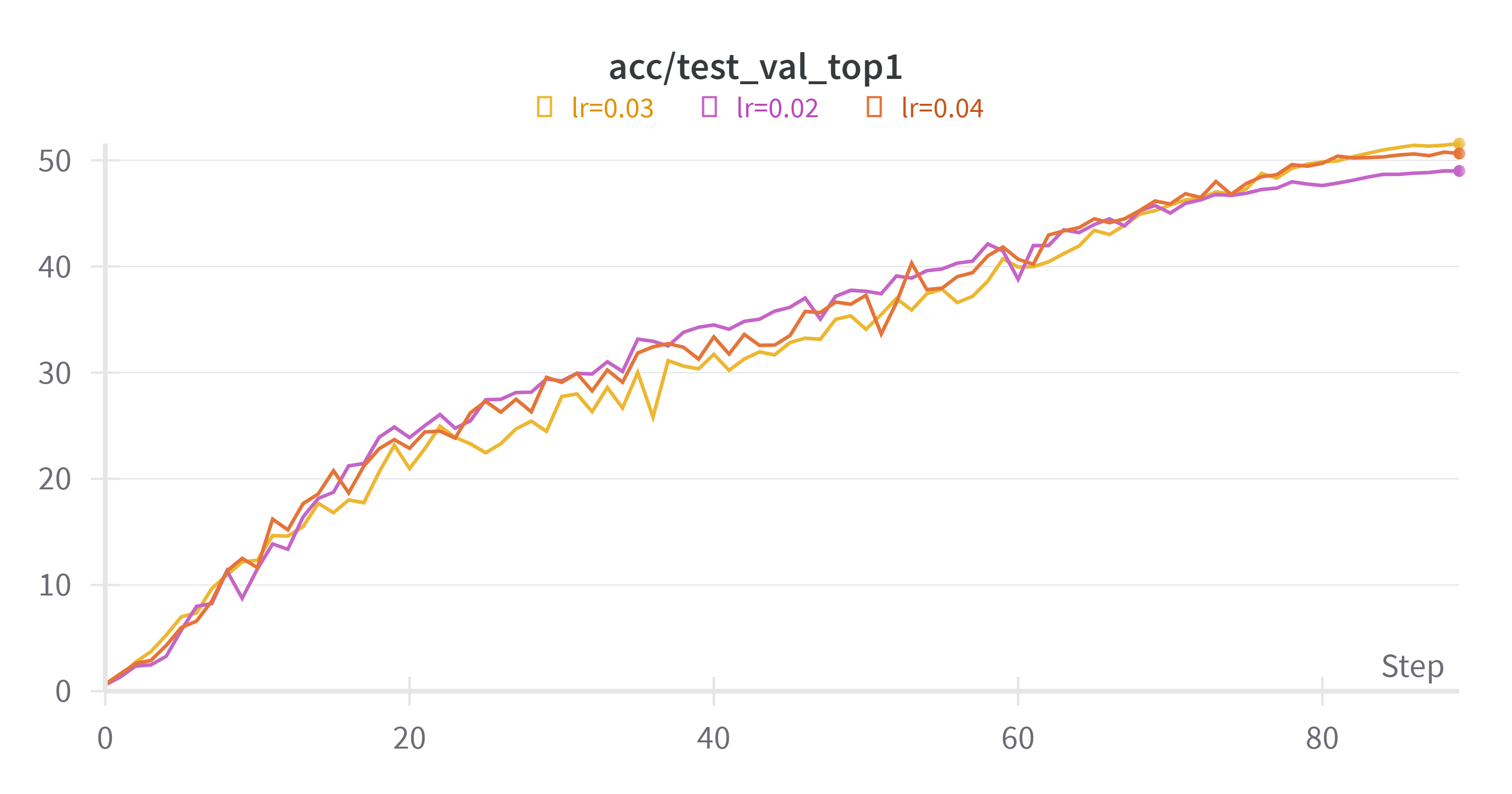}
    \caption{Validation accuracy comparison for different initial learning rates (0.02, 0.03, 0.04) using Cosine Annealing.}
    \label{fig:lr}
\end{figure}

Using a batch size of 128 and the Cosine Annealing learning rate strategy, an initial rate of 0.03 provided the best results, shown in Figure \ref{fig:lr}.

\begin{figure}[htbp]
    \centering
    \includegraphics[width=\columnwidth]{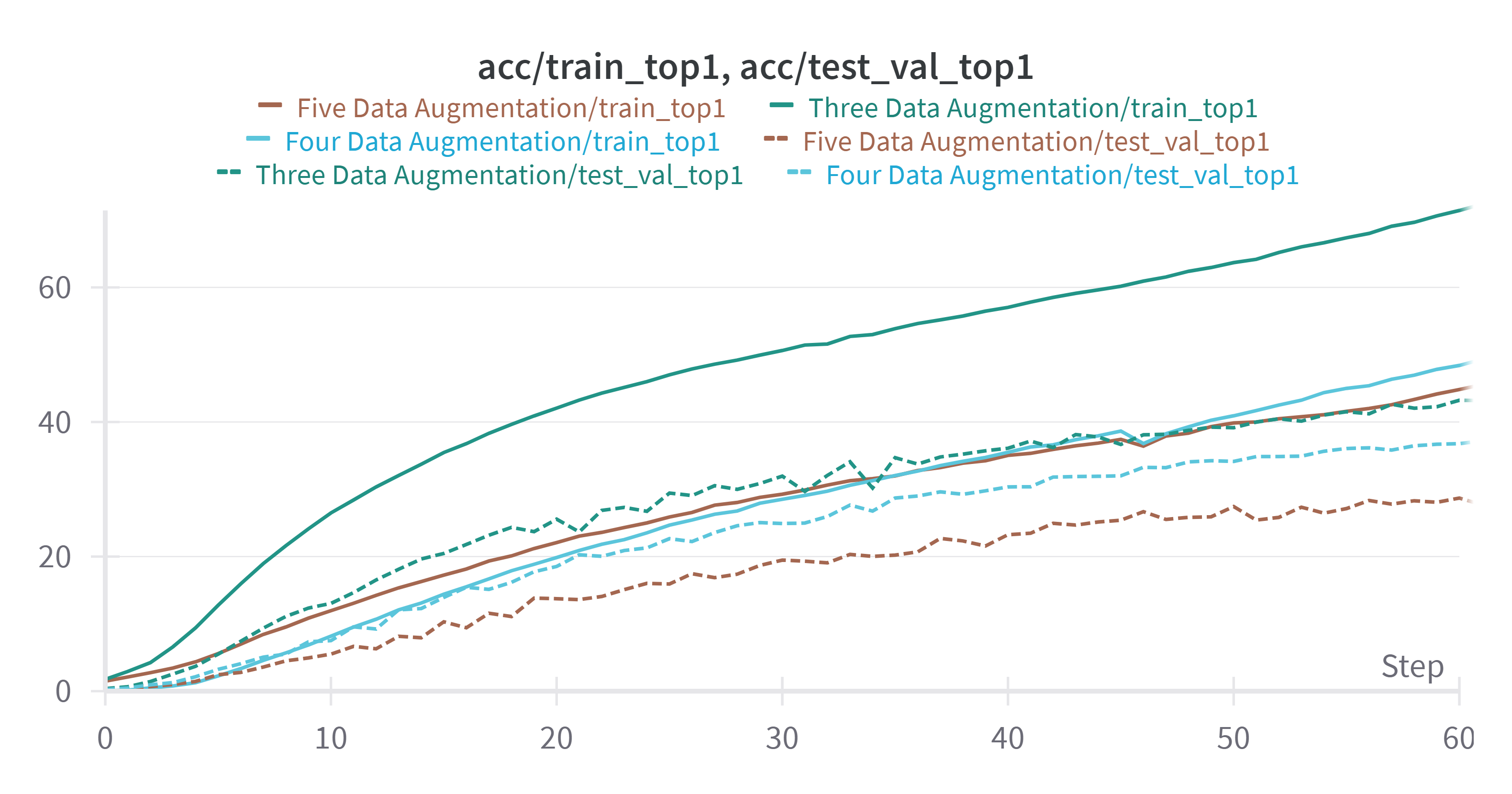}
    \caption{Accuracy for training and validation sets with three, four, and five data augmentation techniques. 'Three' includes random size cropping, random horizontal flipping, and color jittering; 'Four' adds random grayscale; 'Five' incorporates all the previous techniques plus random affine.}
    \label{fig:data_augmentation}
\end{figure}

In our final experiments, we used three data augmentation techniques: random size cropping, random horizontal flipping, and color jittering. Additionally, we experimented with adding random grayscale and random affine with parameters degrees=(10, 150), translate=(0.1,0.6), and shear=45. According to Figure \ref{fig:data_augmentation}, while more data augmentation techniques did reduce the difference between training and validation accuracy, they also lowered accuracy in both training and validation sets. We believe that excessive data augmentation led to greater dispersion among samples of the same class, which contradicts our objective of clustering samples within classes. Because of noting the adverse effects of extensive data augmentation, we ceased training earlier; hence, Figure \ref{fig:data_augmentation} only displays results up to 60 epochs.

\begin{figure}[htbp]
    \centering
    \includegraphics[width=\columnwidth]{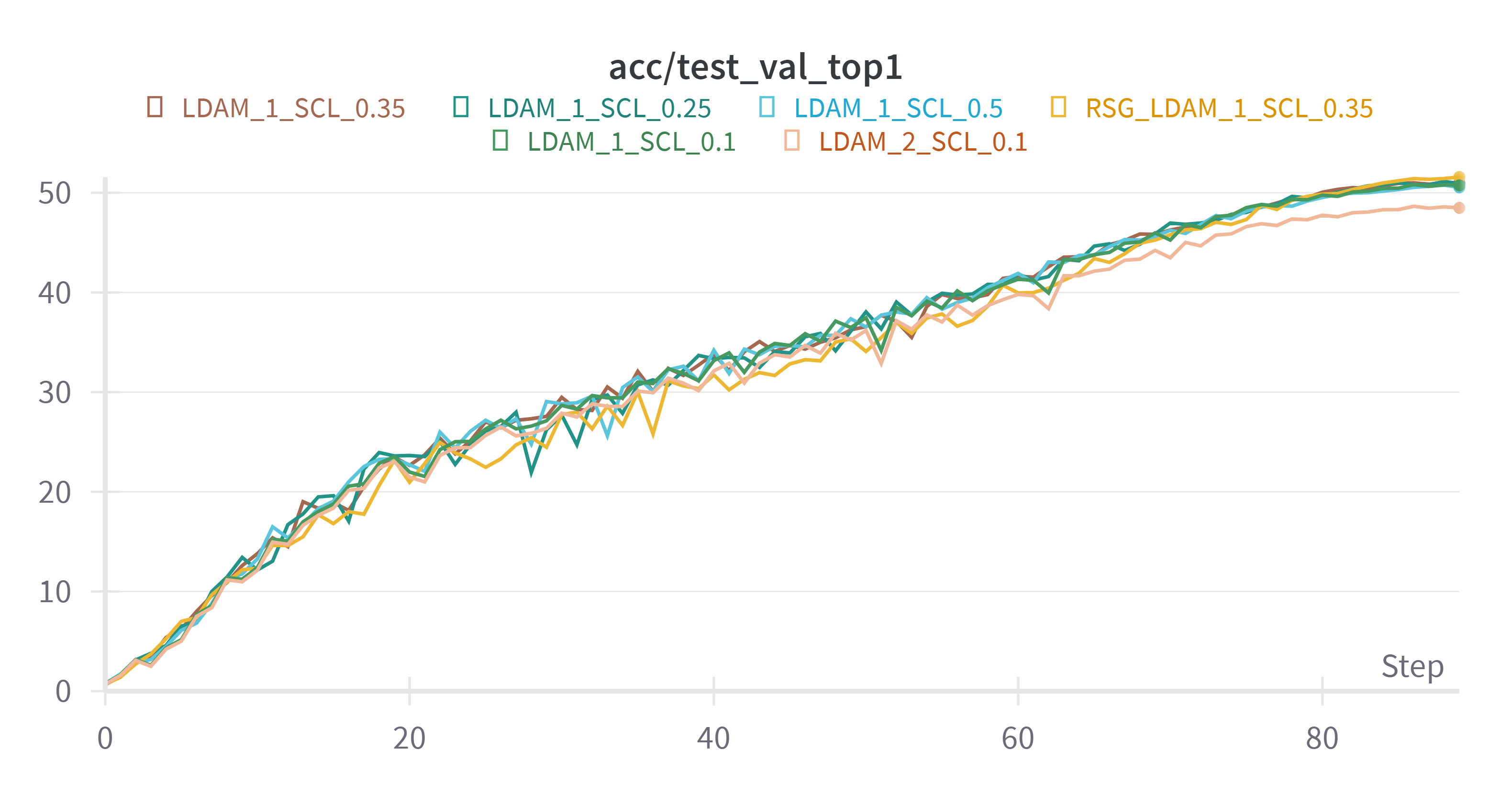}
    \caption{Validation accuracy of SCL-LDAM on the ImageNet-LT dataset with several loss weights. Observation suggests that setting the SCL weight to 0.35 and LDAM weight to 1 yields better results compared to other weight combinations. Thus, validation accuracy of RSG-SCL-LDAM was also tested using the weight combination.}
    \label{fig:T-results}
\end{figure}

Due to the larger sample size of the ImageNet-LT dataset, using a genetic algorithm was time-consuming. We adopted weights for SCL and LDAM based on \cite{Balanced_contrastive} and conducted several simple weight adjustment tests. The results across 90 epochs are shown in Figure \ref{fig:T-results}.

\section*{Rationale}
\label{sec:rationale}
Previous research on long-tail recognition (LTR) has achieved high accuracy, often at the expense of dominant classes, to improve the accuracy of tail classes. Such methods are advantageous for applications like medical diagnosis and fraud detection. However, misclassifying dominant classes is detrimental in areas like email filtering and recommendation systems. Therefore, our research aims to balance the accuracy improvement across all classes. Our method is simple, making it suitable for LTR problems that require balanced accuracy enhancement. 

The \cite{Hybrid_contrastive} applies two-stage training of Supervised Contrastive Learning (SCL) to end-to-end models, demonstrating the advantage of combining SCL with Cross-Entropy Loss (CE) in LTR. The \cite{Balanced_contrastive} improves SCL to BCL and combines BCL with logits adjustment loss function \cite{logit_adjustment}, achieving higher and more balanced accuracy improvements. A research gap exists between these two papers, lacking an analysis of the contribution of SCL and logits adjustment loss to LTR problems. In our study, we replace CE in the combination of SCL and CE from \cite{Hybrid_contrastive} with Label-Distribution-Aware Margin Loss (LDAM) \cite{LDAM} from the logits adjustment loss to analyze whether logits adjustment contributes to the accuracy of the combination. The SCL and LDAM combination results also provide a baseline for the results in \cite{Balanced_contrastive}, offering a more intuitive understanding of the contribution of balanced SCL compared to standard SCL in LTR problems. Furthermore, we tested the effect of model intagerating the Rare-Class Sample Generator \cite{RSG}. The results show that effectively leveraging the strengths of different LTR techniques can enable them to work collaboratively for better and more balanced outcomes.



\clearpage
{
    \small
    \bibliographystyle{ieeenat_fullname}
    \bibliography{main}
}


\end{document}